\documentclass[10pt,journal,compsoc]{IEEEtran}
\usepackage{times}
\usepackage{graphicx}
\usepackage{amsmath}
\usepackage{bm}
\usepackage{multirow}
\usepackage{amssymb}
\usepackage{amsthm}
\usepackage{subfigure}
\usepackage{url}
\usepackage{wrapfig}
\usepackage{multirow}
\usepackage{algorithm}
\usepackage{algpseudocode}
\usepackage{paralist}
\usepackage{hyperref}

\ifCLASSOPTIONcompsoc
  \usepackage[nocompress]{cite}
\else
  \usepackage{cite}
\fi
\ifCLASSINFOpdf
\else
\fi
\begin{document}
%
% paper title
% Titles are generally capitalized except for words such as a, an, and, as,
% at, but, by, for, in, nor, of, on, or, the, to and up, which are usually
% not capitalized unless they are the first or last word of the title.
% Linebreaks \\ can be used within to get better formatting as desired.
% Do not put math or special symbols in the title.
\title{A Survey on Multi-Task Learning}
%
%
% author names and IEEE memberships
% note positions of commas and nonbreaking spaces ( ~ ) LaTeX will not break
% a structure at a ~ so this keeps an author's name from being broken across
% two lines.
% use \thanks{} to gain access to the first footnote area
% a separate \thanks must be used for each paragraph as LaTeX2e's \thanks
% was not built to handle multiple paragraphs
%
%
%\IEEEcompsocitemizethanks is a special \thanks that produces the bulleted
% lists the Computer Society journals use for "first footnote" author
% affiliations. Use \IEEEcompsocthanksitem which works much like \item
% for each affiliation group. When not in compsoc mode,
% \IEEEcompsocitemizethanks becomes like \thanks and
% \IEEEcompsocthanksitem becomes a line break with idention. This
% facilitates dual compilation, although admittedly the differences in the
% desired content of \author between the different types of papers makes a
% one-size-fits-all approach a daunting prospect. For instance, compsoc
% journal papers have the author affiliations above the "Manuscript
% received ..."  text while in non-compsoc journals this is reversed. Sigh.

\author{Yu~Zhang and Qiang Yang% <-this % stops a space
\IEEEcompsocitemizethanks{\IEEEcompsocthanksitem Y. Zhang is with Department of Computer Science and Engineering, Southern University of Science and Technology and Peng Cheng Laboratory. Q. Yang is with the Department of Computer Science and Engineering, Hong Kong University of Science and Technology.\protect\\
E-mail: yu.zhang.ust@gmail.com, qyang@cse.ust.hk\protect\\
Corresponding author: Yu Zhang.}
% <-this % stops a space
%\thanks{Manuscript received February, 2014; revised November, 2014.}
}

% note the % following the last \IEEEmembership and also \thanks -
% these prevent an unwanted space from occurring between the last author name
% and the end of the author line. i.e., if you had this:
%
% \author{....lastname \thanks{...} \thanks{...} }
%                     ^------------^------------^----Do not want these spaces!
%
% a space would be appended to the last name and could cause every name on that
% line to be shifted left slightly. This is one of those "LaTeX things". For
% instance, "\textbf{A} \textbf{B}" will typeset as "A B" not "AB". To get
% "AB" then you have to do: "\textbf{A}\textbf{B}"
% \thanks is no different in this regard, so shield the last } of each \thanks
% that ends a line with a % and do not let a space in before the next \thanks.
% Spaces after \IEEEmembership other than the last one are OK (and needed) as
% you are supposed to have spaces between the names. For what it is worth,
% this is a minor point as most people would not even notice if the said evil
% space somehow managed to creep in.

\newtheorem{theorem}{Theorem}
\newtheorem{lemma}{Lemma}
\newtheorem*{definition}{Definition}

\newcommand{\tr}{\mathrm{tr}}		% trace
\newcommand{\I}{\mathbf{I}}        % an identity matrix
\newcommand{\0}{\mathbf{0}}       % a zero vector or matrix
\newcommand{\1}{\mathbf{1}}       % a vector or matrix of all ones

\newcommand{\E}{\mathbb{E}} 		% expectaion
\newcommand{\p}{\mathbb{P}}		% probablity

\newcommand{\K}{\mathbf{K}} 	% kernel matrix

\newcommand{\kl}{\mathrm{D_{KL}}}		% K-L divergence
\newcommand{\breg}{\mathrm{D_{Breg}}} 	% Bregmen divergence

\newcommand{\h}{\mathcal{H}}	 % hypothesis space
\newcommand{\err}{\epsilon}		 % error
\newcommand{\emerr}{\hat{\epsilon}} % empirical error

\newcommand{\ds}{\mathcal{D}}	% a dataset
\newcommand{\fea}{\mathcal{X}}	% feature
\newcommand{\dm}{\mathbf{X}}		% data matrix
\newcommand{\lv}{\mathbf{y}}		% label vector
\newcommand{\soc}{\mathbb{S}}	% the source task
\newcommand{\tar}{\mathbb{T}}	% the target task
\newcommand{\tarx}{\mathbf{X}_t}	% features for target task
\newcommand{\tary}{\mathbf{y}_t}		% label for target task

\newcommand{\w}{\mathbf{w}} 			% coefficient vector
\newcommand{\W}{\mathbf{W}} 			% coefficient matrix

\newcommand{\x}{\mathbf{x}}

\def\etal{{\em et al.\/}\,}

% The paper headers
\markboth{IEEE TRANSACTIONS on IEEE TRANSACTIONS ON KNOWLEDGE AND DATA ENGINEERING}{Y. Zhang and Q. Yang: A Survey on Multi-Task Learning}
% The only time the second header will appear is for the odd numbered pages
% after the title page when using the twoside option.
%
% *** Note that you probably will NOT want to include the author's ***
% *** name in the headers of peer review papers.                   ***
% You can use \ifCLASSOPTIONpeerreview for conditional compilation here if
% you desire.

% The publisher's ID mark at the bottom of the page is less important with
% Computer Society journal papers as those publications place the marks
% outside of the main text columns and, therefore, unlike regular IEEE
% journals, the available text space is not reduced by their presence.
% If you want to put a publisher's ID mark on the page you can do it like
% this:
%\IEEEpubid{0000--0000/00\$00.00~\copyright~2014 IEEE}
% or like this to get the Computer Society new two part style.
%\IEEEpubid{\makebox[\columnwidth]{\hfill 0000--0000/00/\$00.00~\copyright~2014 IEEE}%
%\hspace{\columnsep}\makebox[\columnwidth]{Published by the IEEE Computer Society\hfill}}
% Remember, if you use this you must call \IEEEpubidadjcol in the second
% column for its text to clear the IEEEpubid mark (Computer Society journal
% papers don't need this extra clearance.)

% use for special paper notices
%\IEEEspecialpapernotice{(Invited Paper)}

% for Computer Society papers, we must declare the abstract and index terms
% PRIOR to the title within the \IEEEtitleabstractindextext IEEEtran
% command as these need to go into the title area created by \maketitle.
% As a general rule, do not put math, special symbols or citations
% in the abstract or keywords.
\IEEEtitleabstractindextext{%
\begin{abstract}

Multi-Task Learning (MTL) is a learning paradigm in machine learning and its aim is to leverage useful information contained in multiple related tasks to help improve the generalization performance of all the tasks. In this paper, we give a survey for MTL from the perspective of algorithmic modeling, applications and theoretical analyses. For algorithmic modeling, we give a definition of MTL and then classify different MTL algorithms into five categories, including feature learning approach, low-rank approach, task clustering approach, task relation learning approach and decomposition approach as well as discussing the characteristics of each approach. In order to improve the performance of learning tasks further, MTL can be combined with other learning paradigms including semi-supervised learning, active learning, unsupervised learning, reinforcement learning, multi-view learning and graphical models. When the number of tasks is large or the data dimensionality is high, we review online, parallel and distributed MTL models as well as dimensionality reduction and feature hashing to reveal their computational and storage advantages. Many real-world applications use MTL to boost their performance and we review representative works in this paper. Finally, we present theoretical analyses and discuss several future directions for MTL.

\end{abstract}

% Note that keywords are not normally used for peerreview papers.
}

% make the title area
\maketitle

% To allow for easy dual compilation without having to reenter the
% abstract/keywords data, the \IEEEtitleabstractindextext text will
% not be used in maketitle, but will appear (i.e., to be "transported")
% here as \IEEEdisplaynontitleabstractindextext when compsoc mode
% is not selected <OR> if conference mode is selected - because compsoc
% conference papers position the abstract like regular (non-compsoc)
% papers do!
\IEEEdisplaynontitleabstractindextext
% \IEEEdisplaynontitleabstractindextext has no effect when using
% compsoc under a non-conference mode.

% For peer review papers, you can put extra information on the cover
% page as needed:
% \ifCLASSOPTIONpeerreview
% \begin{center} \bfseries EDICS Category: 3-BBND \end{center}
% \fi
%
% For peerreview papers, this IEEEtran command inserts a page break and
% creates the second title. It will be ignored for other modes.
\IEEEpeerreviewmaketitle

\ifCLASSOPTIONcompsoc
\IEEEraisesectionheading{\section{Introduction}\label{sec:introduction}}
\else
\section{Introduction}
\label{sec:intro}
\fi
% Computer Society journal (but not conference!) papers do something unusual
% with the very first section heading (almost always called "Introduction").
% They place it ABOVE the main text! IEEEtran.cls does not automatically do
% this for you, but you can achieve this effect with the provided
% \IEEEraisesectionheading{} command. Note the need to keep any \label that
% is to refer to the section immediately after \section in the above as
% \IEEEraisesectionheading puts \section within a raised box.

% The very first letter is a 2 line initial drop letter followed
% by the rest of the first word in caps (small caps for compsoc).
%
% form to use if the first word consists of a single letter:
% \IEEEPARstart{A}{demo} file is ....
%
% form to use if you need the single drop letter followed by
% normal text (unknown if ever used by IEEE):
% \IEEEPARstart{A}{}demo file is ....
%
% Some journals put the first two words in caps:
% \IEEEPARstart{T}{his demo} file is ....
%
% Here we have the typical use of a "T" for an initial drop letter
% and "HIS" in caps to complete the first word.

\IEEEPARstart{H}{uman} can learn multiple tasks simultaneously and during this learning process, human can use the knowledge learned in a task to help the learning of another task. For example, according to our experience in learning to play tennis and squash, we find that the skill of playing tennis can help learn to play squash and vice versa. Inspired by such human learning ability, Multi-Task Learning (MTL) \cite{caruana97}, a learning paradigm in machine learning, aims to learn multiple related tasks jointly so that the knowledge contained in a task can be leveraged by other tasks, with the hope of improving the generalization performance of all the tasks at hand.

At its early stage, an important motivation of MTL is to alleviate the data sparsity problem where each task has a limited number of labeled data. In the data sparsity problem, the number of labeled data in each task is insufficient to train an accurate learner, while MTL aggregates the labeled data in all the tasks in the spirit of data augmentation to obtain a more accurate learner for each task. From this perspective, MTL can help reuse existing knowledge and reduce the cost of manual labeling for learning tasks. When the era of ``big data'' comes in some areas such as computer vision and Natural Language Processing (NLP), it is found that deep MTL models can achieve better performance than their single-task counterparts. One reason that MTL is effective is that it utilizes more data from different learning tasks when compared with single-task learning. With more data, MTL can learn more robust and universal representations for multiple tasks and more powerful models, leading to better knowledge sharing among tasks, better performance of each task and low risk of overfitting in each task.

\begin{figure*}[!htb]
\centering
\subfigure[MTL vs. transfer learning]{
\includegraphics[width=0.26\textwidth]{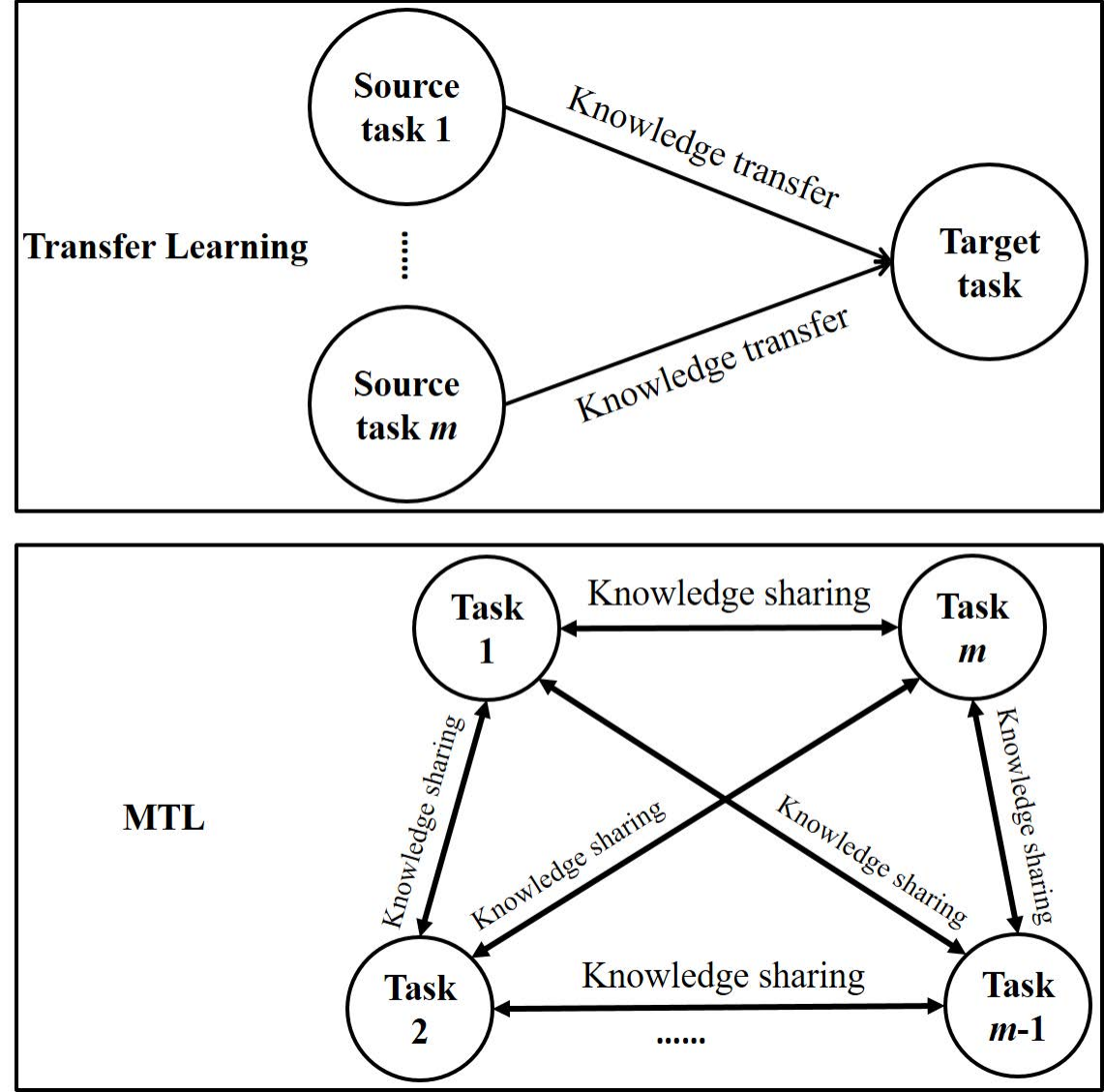}
\label{fig:diff_TL}
}
\hspace{0.15in}
\subfigure[MTL vs. multi-label learning/multi-output regression]{
\includegraphics[width=0.246\textwidth]{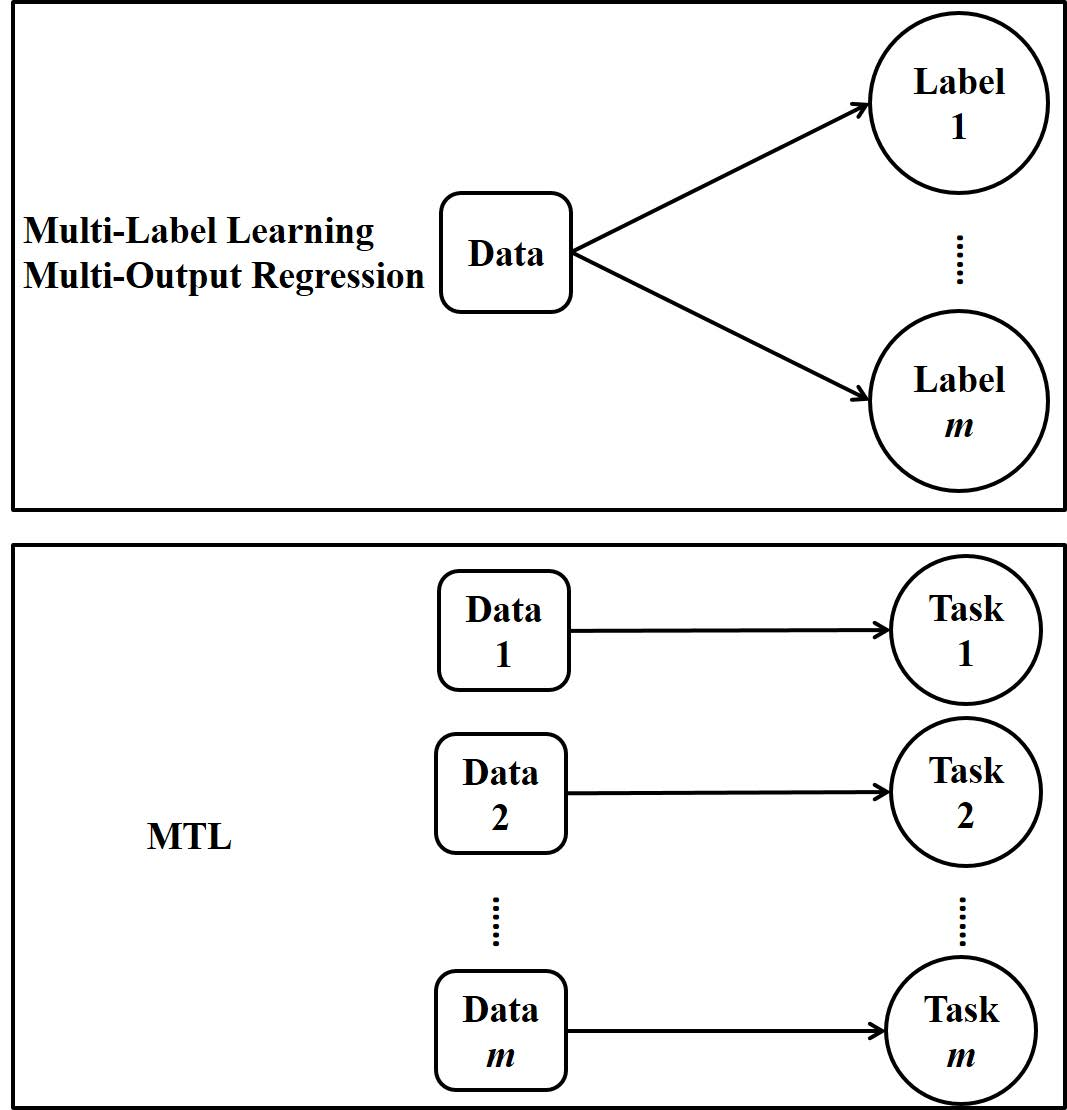}
\label{fig:diff_MLL}
}
\hspace{0.15in}
\subfigure[MTL vs. multi-view learning]{
\includegraphics[width=0.28\textwidth]{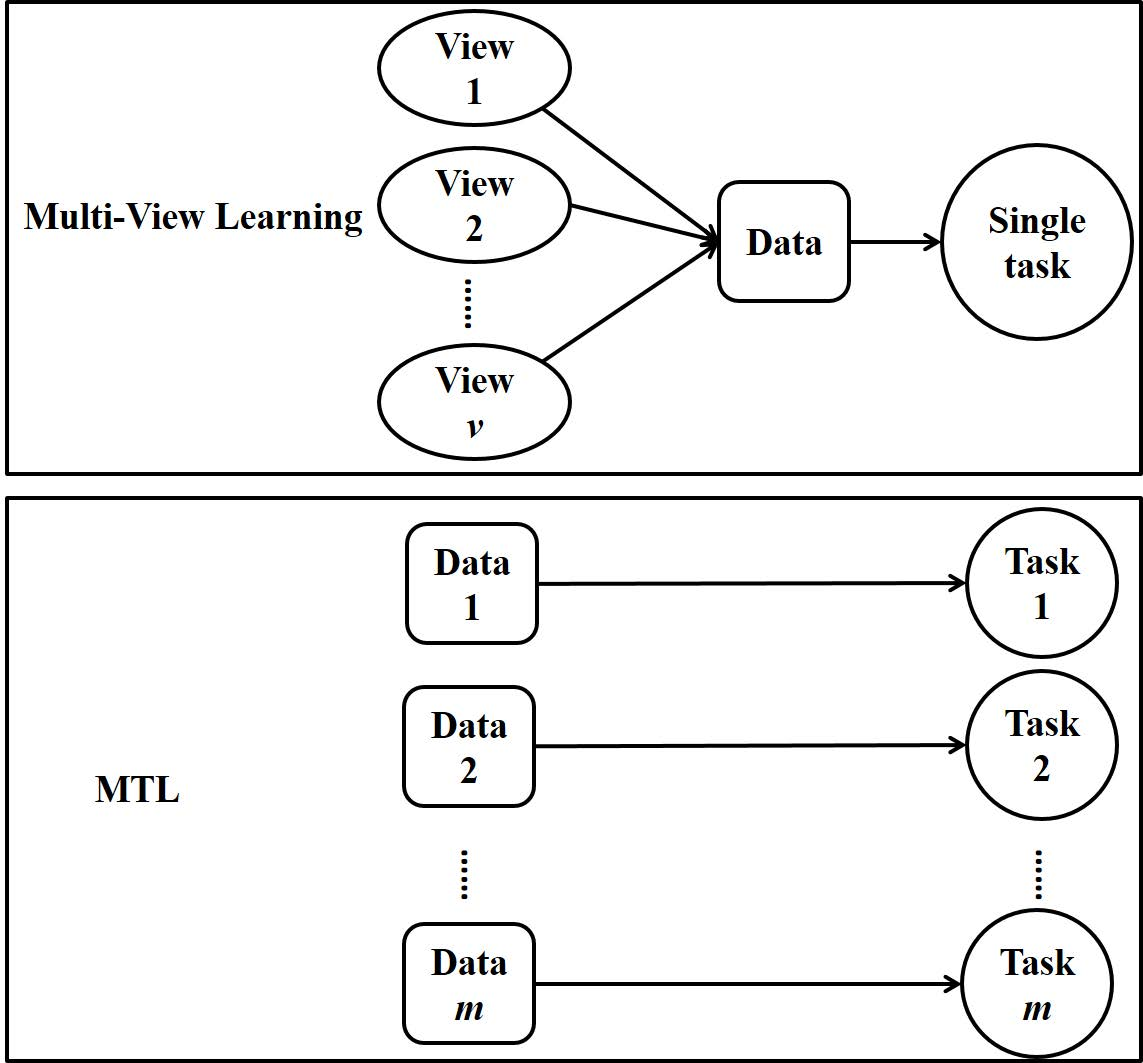}
\label{fig:diff_MVL}
}
\caption{Illustrations for differences between MTL and other learning paradigms.}
\label{fig:diff}
\end{figure*}

MTL is related to other learning paradigms in machine learning, including transfer learning \cite{yzdp20}, multi-label learning \cite{zz14} and multi-output regression. The setting of MTL is similar to that of transfer learning but with significant differences. In MTL, there is no distinction among different tasks and the objective is to improve the performance of all the tasks. However, transfer learning is to improve the performance of a target task with the help of source tasks, hence the target task plays a more important role than source tasks. In a word, MTL treats all the tasks equally but in transfer learning the target task attracts most attentions. From the perspective of the knowledge flow, flows of knowledge transfer in transfer learning are from source task(s) to the target task, but in multi-task learning, there are flows of knowledge sharing between any pair of tasks, which is illustrated in Fig. \ref{fig:diff_TL}. Continual learning \cite{pkpkw19}, in which tasks come sequentially, learns tasks one by one, while MTL is to learn multiple tasks together. In multi-label learning and multi-output regression, each data point is associated with multiple labels which can be categorical or numeric. If we treat each of all the possible labels as a task, multi-label learning and multi-output regression can be viewed in some sense as a special case of multi-task learning where different tasks always share the same data during both the training and testing phrases. On the one hand, such characteristic in multi-label learning and multi-output regression leads to different research issues from MTL. For example, the ranking loss, which enforces the scores (e.g., the classification probability) of labels associated with a data point to be larger than those of absent labels, can be used for multi-label learning but it does not fit MTL where different tasks possess different data. On the other hand, this characteristic in multi-label learning and multi-output regression is invalid in MTL problems. For example, in a MTL problem discussed in Section \ref{sec:dataset} where each task is to predict the disease symptom score of Parkinson for a patient based on 19 bio-medical features, different patients/tasks should not share the bio-medical data. In a word, multi-label learning and multi-output regression are different from multi-task learning as illustrated in Fig. \ref{fig:diff_MLL} and hence we will not survey literature on multi-label learning and multi-output regression. Moreover, multi-view learning is another learning paradigm in machine learning, where each data point is associated with multiple views, each of which consists of a set of features. Even though different views have different sets of features, all the views are used together to learn for the same task and hence multi-view learning belongs to single-task learning with multiple sets of features, which is different from MTL as shown in Fig. \ref{fig:diff_MVL}.

Over past decades, MTL has attracted many attentions in the artificial intelligence and machine learning communities. Many MTL models have been devised and many MTL applications in other areas have been exploited. Moreover, many analyses have been conducted to study theoretical problems in MTL. This paper serves as a survey on MTL from the perspective of algorithmic modeling, applications and theoretical analyses. For algorithmic modeling, we first give a definition for MTL and then classify different MTL algorithms into five categories: feature learning approach which can be further categorized into feature transformation and feature selection approaches, low-rank approach, task clustering approach, task relation learning approach and decomposition approach. After that, we discuss the combination of MTL with other learning paradigms, including semi-supervised learning, active learning, unsupervised learning, reinforcement learning, multi-view learning and graphical models. To handle a large number of tasks, we review online, parallel and distributed MTL models. For data in a high-dimensional space, feature selection, dimensionality reduction and feature hashing are introduced as vital tools to process them. As a promising learning paradigm, MTL has many applications in various areas and here we briefly review its applications in computer vision, bioinformatics, health informatics, speech, NLP, web, etc. From the perspective of theoretical analyses on MTL, we review relevant works. At last, we discuss several future directions for MTL.\footnote{For an introduction to MTL without technical details, please refer to \cite{zy18}.}

\section{MTL Models}
\label{sec:category}

In order to fully characterize MTL, we first give the definition of MTL.

\begin{definition}[Multi-Task Learning] Given $m$ learning tasks $\{\mathcal{T}_i\}_{i=1}^m$ where all the tasks or a subset of them are related, multi-task learning aims to learn the $m$ tasks together to improve the learning of a model for each task $\mathcal{T}_i$ by using the knowledge contained in all or some of other tasks.
\end{definition}

Based on the definition of MTL, we focus on supervised learning tasks in this section since most MTL studies fall in this setting and for other types of tasks, we review them in the next section. In the setting of supervised learning tasks, a task $\mathcal{T}_i$ is usually accompanied by a training dataset $\ds_i$ consisting of $n_i$ training samples, i.e., $\ds_i=\{\mathbf{x}^i_j,y^i_j\}_{j=1}^{n_i}$, where $\mathbf{x}^i_j\in\mathbb{R}^{d_i}$ is the $j$th training instance in $\mathcal{T}_i$ and $y^i_j$ is its label. We denote by $\mathbf{X}^i$ the training data matrix for $\mathcal{T}_i$, i.e., $\mathbf{X}^i=(\mathbf{x}^i_1,\ldots,\mathbf{x}^i_{n_i})$. When different tasks lie in the same feature space implying that $d_i$ equals $d_j$ for any $i\ne j$, this setting is the homogeneous-feature MTL, and otherwise it corresponds to heterogeneous-feature MTL. Without special explanation, the default MTL setting is the homogeneous-feature MTL. Here we need to distinguish the heterogeneous-feature MTL from the heterogeneous MTL. In \cite{ykx09}, the heterogeneous MTL is considered to consist of different types of supervised tasks including classification and regression problems, and here we generalize it to a more general setting that the heterogeneous MTL consists of tasks with different types including supervised learning, unsupervised learning, semi-supervised learning, reinforcement learning, multi-view learning and graphical models. The opposite to the heterogeneous MTL is the homogeneous MTL which consists of tasks with only one type. In a word, the homogeneous and heterogeneous MTL differ in the type of learning tasks while the homogeneous-feature MTL is different from the heterogeneous-feature MTL in terms of the original feature representations. Similarly, without special explanation, the default MTL setting is the homogeneous MTL.

In order to characterize the relatedness in the definition of MTL, there are three issues to be addressed: when to share, what to share and how to share.

The `when to share' issue is to make choices between single-task and multi-task models for a multi-task problem. Currently such decision is made by human experts and there are few works to study it. A simple solution is to formulate such decision as a model selection problem and then use model selection techniques, e.g., cross validation, to make decisions, but this solution is usually computational heavy and may require much more training data. An advanced solution we think is to use multi-task models which can degenerate to their single-task counterparts given some form of model parameters, for example, problem (\ref{equ_second_category_obj}) presented in Section \ref{sec:another_taxonomy} which can reduce to multiple single-task models with the learning of different tasks decoupled when a parameter $\bm{\Sigma}$ becomes diagonal. In this case, we can let the training data determine the form of $\bm{\Sigma}$ to make an implicit choice.

`What to share' needs to determine the form through which knowledge sharing among all the tasks could occur. Usually, there are three forms for `what to share', including feature, instance and parameter. Feature-based MTL aims to learn common features among different tasks as a way to share knowledge. Instance-based MTL identifies useful data instances in a task for other tasks and then shares knowledge via the identified instances. Parameter-based MTL uses model parameters (e.g., coefficients in linear models or weights in deep models) in a task to help learn model parameters in other tasks in some ways, for example, the regularization. Existing MTL studies mainly focus on feature-based and parameter-based methods, and only a few works belong to the instance-based method. A representative instance-based method is the multi-task distribution matching method proposed in \cite{bbls08}, which first estimates density ratios between probabilities that each instance as well as its label belongs to both its own task and a mixture of all the tasks and then uses all the weighted training data from all the tasks based on the estimated density ratios to learn model parameters for each task. Since the studies on instance-based MTL are few, we mainly review feature-based and parameter-based MTL models.

After determining `what to share', `how to share' specifies concrete ways to share knowledge among tasks. In feature-based MTL, there is a primary approach: feature learning approach. The feature learning approach focuses on learning common feature representations for multiple tasks based on shallow or deep models, where the learned common feature representation can be a subset or a transformation of the original feature representation.
In parameter-based MTL, there are four main approaches: low-rank approach, task clustering approach, task relation learning approach and decomposition approach. The low-rank approach interprets the relatedness of multiple tasks as the low rankness of the parameter matrix of these tasks. The task clustering approach is to identify task clusters, each of which contains similar tasks. The task relation learning approach aims to learn quantitative relations between tasks from data automatically. The decomposition approach decomposes the model parameters of all the tasks into two or more components, which are penalized by different regularizers.

In summary, there are mainly five approaches in the feature-based and parameter-based MTL. In the following sections, we review these approaches in a chronological order to reveal the relations and evolutions among different models.

\subsection{Feature Learning Approach}

Since tasks are related, it is intuitive to assume that different tasks share a common feature representation based on the original features. One reason to learn common feature representations instead of directly using the original ones is that the original representation may not have enough expressive power for multiple tasks. With the training data in all the tasks, a more powerful representation can be learned for all the tasks and this representation can bring the improvement on the performance.

Based on the relationship between the original feature representation and the learned one, we can further classify this category into two sub-categories. The first sub-category is the feature transformation approach where the learned representation is a linear or nonlinear transformation of the original representation and in this approach, each feature in the learned representation is different from the original features. Different from this approach, the feature selection approach, the second sub-category, selects a subset of the original features as the learned representation and hence the learned representation is similar to the original one by eliminating useless features based on different criteria. In the following, we introduce these two approaches.

\subsubsection{Feature Transformation Approach}

\begin{figure}[!ht]
\centering
\includegraphics[width=.33\textwidth]{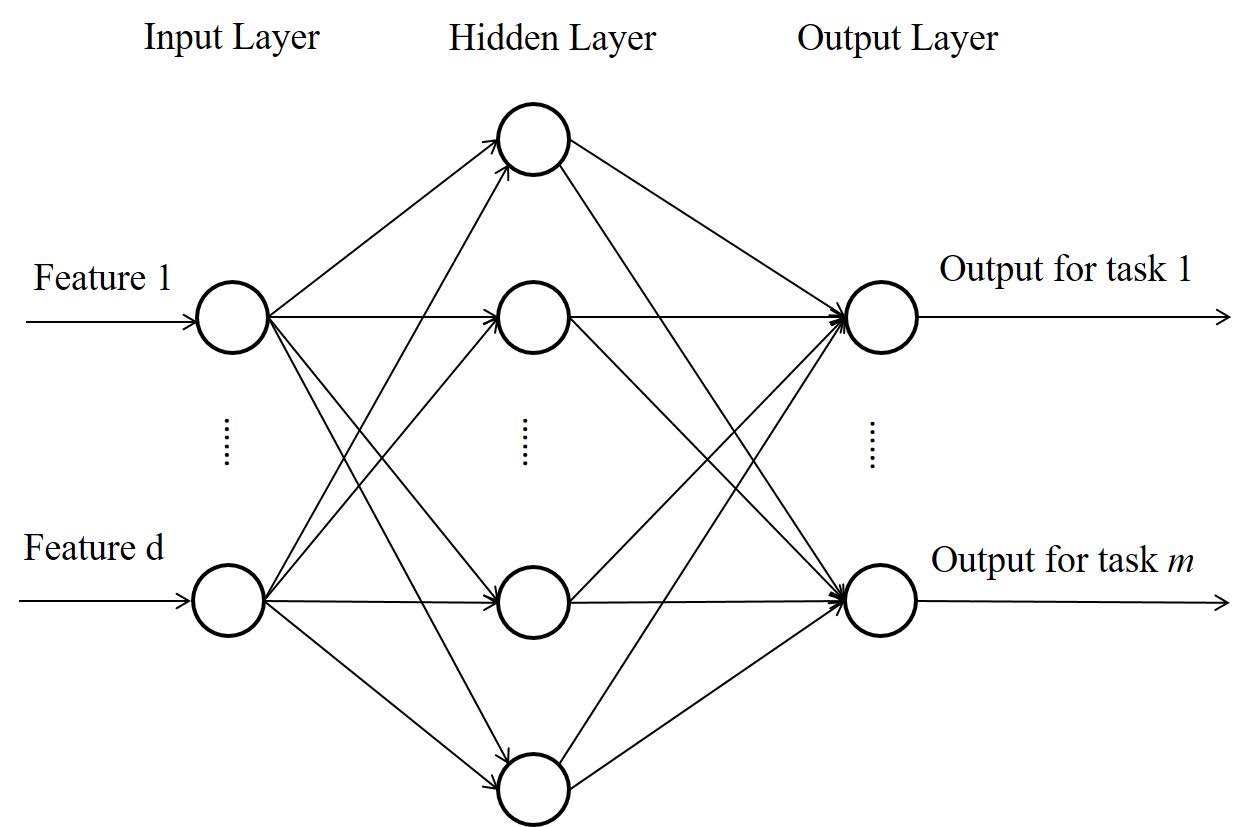}
\caption{An example for the multi-task feedforward neural network with an input layer, a hidden layer and an output layer.
}
\label{fig_MTLNN_sample}
\end{figure}

The multi-layer feedforward neural network~\cite{caruana97}, which belongs to the feature transformation approach, is one of the earliest model for multi-task learning. To see how the multi-layer feedforward neural network is constructed for MTL, in Figure \ref{fig_MTLNN_sample} we show an example with an input layer, a hidden layer and an output layer. The input layer receives training instances from all the tasks and the output layer has $m$ output units with one for each task. Here the outputs of the hidden layer can be viewed as the common feature representation learned for the $m$ tasks and the transformation from the original representation to the learned one depends on the weights connecting the input and hidden layers as well as the activation function adopted in the hidden units. Hence, if the activation function in the hidden layer is linear, then the transformation is a linear function and otherwise it is nonlinear. Compared with multi-layer feedforward neural networks used for single-task learning, the difference in the network architecture lies in the output layers where in single-task learning, there is only one output unit while in MTL, there are $m$ ones. In \cite{lc05}, the radial basis function network, which has only one hidden layer, is extended to MTL by greedily determining the structure of the hidden layer. Different from these neural network models, Silver et al. \cite{spc08} propose a context-sensitive multi-task neural network which has only one output unit shared by different tasks but has a task-specific context as an additional input.

Different from multi-layer feedforward neural networks which are connectionist models, the multi-task feature learning (MTFL) method \cite{aep08} is formulated under the regularization framework with the objective function as
{\small
\begin{align}
\min_{\mathbf{A},\mathbf{U},\mathbf{b}}\ &\sum_{i=1}^m\frac{1}{n_i} \sum_{j=1}^{n_i}l(y^i_j,(\mathbf{a}^i)^T\mathbf{U}^T\mathbf{x}^i_j+b_i)+\lambda\|\mathbf{A}\|_{2,1}^2\nonumber\\
\mathrm{s.t.}\ &\mathbf{U}\mathbf{U}^T=\mathbf{I},\label{equ_MTFL_objective}
\end{align}
}\noindent
where $l(\cdot,\cdot)$ denotes a loss function such as the hinge loss or square loss, $\mathbf{b}=(b_1,\ldots,b_m)^T$ is a vector of offsets in all the tasks, $\mathbf{U}\in\mathbb{R}^{d\times d}$ is a square transformation matrix, $\mathbf{A}\in\mathbb{R}^{d\times m}$ contains model parameters of all the tasks with its $i$th column $\mathbf{a}^i$ as model parameters for the $i$th task after the transformation, the $\ell_{2,1}$ norm of a matrix $\mathbf{A}$ denoted by $\|\mathbf{A}\|_{2,1}$ equals the sum of the $\ell_2$ norm of rows in $\mathbf{A}$, $\mathbf{I}$ denotes an identity matrix with an appropriate size, and $\lambda$ is a positive regularization parameter. The first term in the objective function of problem (\ref{equ_MTFL_objective}) measures the empirical loss on the training sets of all the tasks and the second one is to enforce $\mathbf{A}$ to be row-sparse via the $\ell_{2,1}$ norm which is equivalent to selecting features after the transformation, while the constraint enforces $\mathbf{U}$ to be orthogonal. Different from the multi-layer feedforward neural network whose hidden representations may be redundant, the orthogonality of $\mathbf{U}$ can prevent the MTFL method from it. As proved in \cite{aep08}, problem (\ref{equ_MTFL_objective}) is equivalent to
{\small
\begin{align}
\min_{\mathbf{W},\mathbf{D},\mathbf{b}} L(\mathbf{W},\mathbf{b})+\lambda\mathrm{tr}(\mathbf{W}^T\mathbf{D}^{-1}\mathbf{W})\
\mathrm{s.t.}\ \mathbf{D}\succeq\mathbf{0},\ \mathrm{tr}(\mathbf{D})\le 1,\label{equ_MTFL_objective_2}
\end{align}
}\noindent
where $L(\mathbf{W},\mathbf{b})=\sum_{i=1}^m\frac{1}{n_i}\sum_{j=1}^{n_i}l(y^i_j,(\mathbf{w}^i)^T\mathbf{x}^i_j+b_i)$ denotes the total training loss, $\mathrm{tr}(\cdot)$ denotes the trace of a square matrix, $\mathbf{w}^i=\mathbf{U}\mathbf{a}^i$ is the model parameter for $\mathcal{T}_i$, $\mathbf{W}=(\mathbf{w}^1,\ldots,\mathbf{w}^m)$, $\mathbf{0}$ denotes a zero vector or matrix with an appropriate size, $\mathbf{M}^{-1}$ for any square matrix $\mathbf{M}$ denotes its inverse when it is nonsingular or otherwise its pseudo inverse, and $\mathbf{B}\succeq\mathbf{C}$ means that $\mathbf{B}-\mathbf{C}$ is positive semidefinite. Based on this formulation, we can see that the MTFL method is to learn a feature covariance $\mathbf{D}$ for all the tasks, which will be interpreted in Section \ref{sec:another_taxonomy} from a probabilistic perspective. Given $\mathbf{D}$, the learning of different tasks can be decoupled and this can facilitate the parallel computing. When given $\mathbf{W}$, $\mathbf{D}$ has an analytical solution as $\mathbf{D}=(\mathbf{W}^T\mathbf{W})^{\frac{1}{2}}/\mathrm{tr}\left((\mathbf{W}^T\mathbf{W})^{\frac{1}{2}}\right)$ and by plugging this solution into problem (\ref{equ_MTFL_objective_2}), we can see that the regularizer on $\mathbf{W}$ is the squared trace norm. Then Argyriou et al. \cite{ampy07} extend problem (\ref{equ_MTFL_objective_2}) to a general formulation where the second term in the objective function becomes $\lambda\mathrm{tr}(\mathbf{W}^Tf(\mathbf{D})\mathbf{W})$ with $f(\mathbf{D})$ operating on the spectrum of $\mathbf{D}$ and discuss the condition on $f(\cdot)$ to make the whole problem convex.

Similar to the MTFL method, the multi-task sparse coding method \cite{mpr13} is to learn a linear transformation on features with the objective function formulated as
{\small
\begin{align}
\hskip -0.15in \min_{\mathbf{A},\mathbf{U},\mathbf{b}}L(\mathbf{UA},\mathbf{b})\
\mathrm{s.t.}\ \|\mathbf{a}^i\|_1\le \lambda\ \forall i\in[m],\|\mathbf{u}^j\|_2\le 1\ \forall j\in [D],\label{equ_MTSC_objective}
\end{align}
}\noindent
where $\mathbf{a}^i$, the $i$th column of $\mathbf{A}$, contains model parameters of the $i$th task, $\mathbf{u}^j$ is the $j$th column in $\mathbf{U}$, $[c]$ for an integer $c$ denotes a set of integers from 1 to $c$, $\|\cdot\|_1$ denotes the $\ell_1$ norm of a vector or matrix and equals the sum of the absolute value of its entries, and $\|\cdot\|_2$ denotes the $\ell_2$ norm of a vector. Here the transformation $\mathbf{U}\in\mathbb{R}^{d\times D}$ is also called the dictionary in sparse coding and shared by all the tasks. Compared with the MTFL method where $\mathbf{U}$ in problem (\ref{equ_MTFL_objective}) is a $d\times d$ orthogonal matrix, $\mathbf{U}$ in problem (\ref{equ_MTSC_objective}) is overcomplete, which implies that $D$ is larger than $d$, with each column having a bounded $\ell_2$ norm.  Another difference is that in problem (\ref{equ_MTFL_objective}) $\mathbf{A}$ is enforced to be row-sparse but in problem (\ref{equ_MTSC_objective}) it is only sparse via the first constraint. With a similar idea to the multi-task sparse coding method, Zhu et al. \cite{zcx11} propose a multi-task infinite support vector machine via the Indian buffet process and the difference is that in \cite{zcx11} the dictionary is sparse and model parameters are non-sparse. In \cite{tl11}, the spike and slab prior is used to learn sparse model parameters for multi-output regression problems where transformed features are induced by Gaussian processes and shared by different outputs.

\begin{figure}[!ht]
\centering
\includegraphics[width=.4\textwidth]{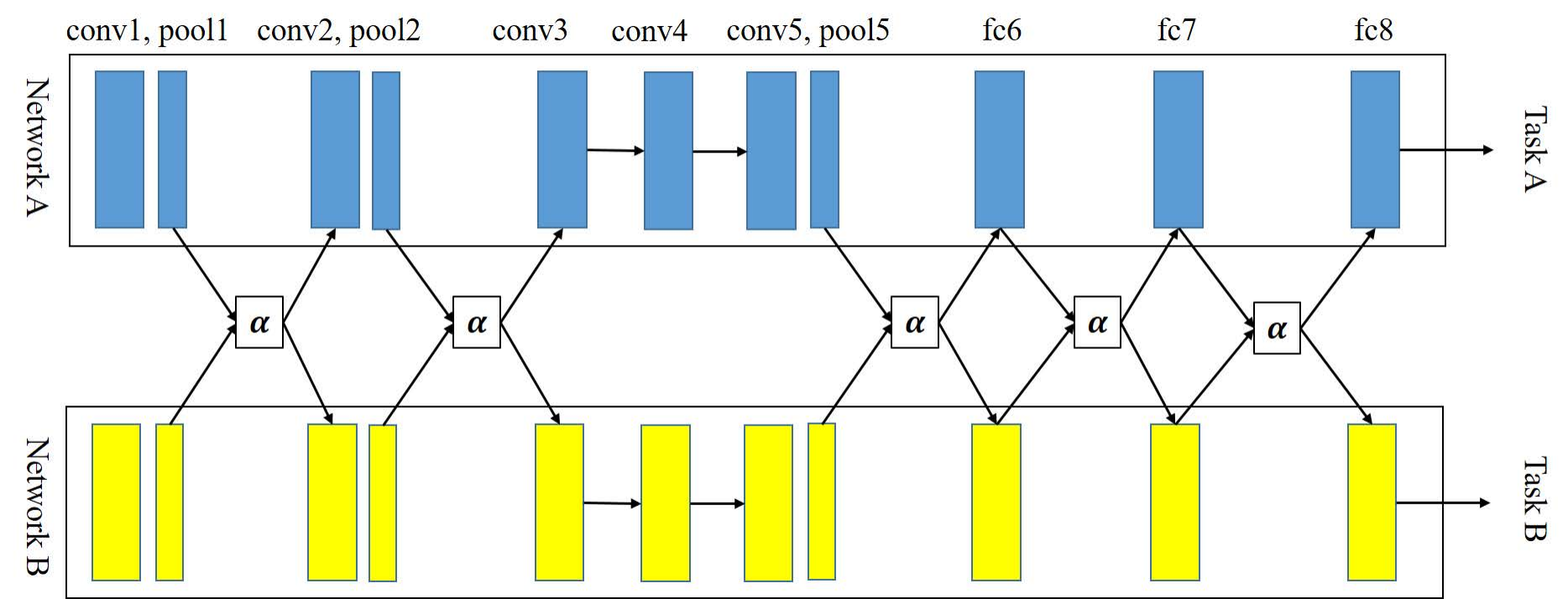}
\caption{The architecture for the cross-stitch network.}
\label{fig_CSN}
\end{figure}

Recently deep learning becomes popular due to its capacity to learn nonlinear features, which facilitates the learning of invariant features for multiple tasks, and hence many deep multi-task models belonging to this approach have been proposed with each task modeled by a deep neural network. Here we classify deep multi-task models in this approach into three main categories. The first category \cite{zllt14,lmzcl15,zlzskyj15,mstgsvwy15,llc15} is to learn a common feature representation for multiple tasks by sharing first several layers in a similar architecture to Fig. \ref{fig_MTLNN_sample}. However, different from Fig. \ref{fig_MTLNN_sample}, deep MTL models in this category have a large number of shared layers, which have general structures such as convolutional layers and pooling layers. Building on the first category, the second category is to use adversarial learning, which is inspired by generative adversarial networks, to learn a common feature representation for MTL as did in \cite{shinohara16,lqh17}. Specifically, there are three networks in such adversarial multi-task models, including a feature network $N_f$, a classification network $N_c$ and a domain network $N_d$. Based on $N_f$, $N_c$ is to minimize the training loss for all the tasks, while $N_d$ aims to distinguish which task a data instance is from. The objective function of such models is usually formulated as
{\small
\begin{equation*}
\min_{\theta_f,\theta_c}\max_{\theta_d} \sum_{i=1}^m\frac{1}{n_i}\sum_{j=1}^{n_i}\left(l(y^i_j,N_c(N_f(\mathbf{x}^i_j)))-l_{ce}(d^i_j,N_d(N_f(\mathbf{x}^i_j)))\right),
\end{equation*}
}\noindent
where $\theta_f$, $\theta_c$, $\theta_d$ denote parameters of three networks $N_f$, $N_c$, $N_d$, respectively, $d^i_j\in\{1,\ldots,m\}$ denotes the task index/label of $\mathbf{x}^i_j$, and $l_{ce}(\cdot,\cdot)$ denotes the cross-entropy loss. Based on this minimax problem, $N_f$ is to minimize the training loss for all the tasks and maximize the cross-entropy loss to fool the domain network to make the learned feature representation indistinguishable to all the tasks. When there is no domain network, this category can reduce to the first category. Moreover, in \cite{lqh17}, each task can learn its specific feature representation to increase the expressive power of the whole model. The last category is to learn different but related feature representations for different tasks with the cross-stitch network \cite{msgh16} as a representative model. Specifically, given two tasks $A$ and $B$ with an identical network architecture, $x^{i,j}_A$ ($x^{i,j}_B$) denotes the hidden feature outputted by the $j$th unit of the $i$th hidden layer for task $A$ ($B$). Then we can define the cross-stitch operation on $x^{i,j}_A$ and $x^{i,j}_B$ as
{\small $\left( \begin{array}{c}
\tilde{x}^{i,j}_A \\
\tilde{x}^{i,j}_B
\end{array} \right)=
\left( \begin{array}{cc}
\alpha_{AA} & \alpha_{AB} \\
\alpha_{BA} & \alpha_{BB}
\end{array} \right)
\left( \begin{array}{ccc}
x^{i,j}_A \\
x^{i,j}_B
\end{array} \right)$},
where $\tilde{x}^{i,j}_A$ and $\tilde{x}^{i,j}_B$ are new hidden features after learning the two tasks jointly. When both $\alpha_{AB}$ and $\alpha_{BA}$ equal 0, training the two networks jointly is equivalent to training them independently. The network architecture of the cross-stitch network is shown in Fig. \ref{fig_CSN}. Here matrix $\boldsymbol{\alpha}$, which is defined as {\small $\boldsymbol{\alpha}\equiv\left( \begin{array}{cc}
\alpha_{AA} & \alpha_{AB} \\
\alpha_{BA} & \alpha_{BB}
\end{array} \right)$},
encodes feature-level task relations between the two tasks and it can be learned via the backpropagation method.

\subsubsection{Feature Selection Approach}

One way to do feature selection in MTL is to use the $\ell_{p,q}$ norm denoted by $\|\mathbf{W}\|_{p,q}\equiv\|(\|\mathbf{w}_1\|_p,\ldots,\|\mathbf{w}_d\|_p)\|_q$, where $\mathbf{w}_i$ denotes the $i$th row of $\mathbf{W}$ and $\|\cdot\|_p$ denotes the $\ell_p$ norm of a vector, to achieve the group sparsity. Obozinski et al. \cite{otj06} are among the first to study the multi-task feature selection (MTFS) problem based on the $\ell_{2,1}$ norm with the objective function formulated as
{\small
\begin{equation}
\min_{\mathbf{W},\mathbf{b}}\ \ L(\mathbf{W},\mathbf{b})+\lambda\|\mathbf{W}\|_{2,1}.\label{equ_MTFS_l21_objective}
\end{equation}
}\noindent
The regularizer on $\mathbf{W}$ in problem (\ref{equ_MTFS_l21_objective}) is to enforce $\mathbf{W}$ to be row-sparse, which in turn helps select important features. In \cite{otj06}, a path-following algorithm is proposed  to solve problem (\ref{equ_MTFS_l21_objective}) and then Liu et al. \cite{ljy09} employ an optimal first-order optimization method to solve it. Compared with problem (\ref{equ_MTFL_objective}), we can see that problem (\ref{equ_MTFS_l21_objective}) is similar to the MTFL method without learning the transformation $\mathbf{U}$. Lee et al. \cite{lzx10} propose a weighted $\ell_{2,1}$ norm for multi-task feature selection where the weights can be learned as well and problem (\ref{equ_MTFS_l21_objective}) is extended in \cite{rcnr13} to a general case where feature groups can overlap with each other. In order to make problem (\ref{equ_MTFS_l21_objective}) more robust to outliers, a square-root loss function is investigated in \cite{gzfy14}. Moreover, in order to make speedup, a safe screening method is proposed in \cite{wy15} to filter out useless features corresponding to zero rows in $\mathbf{W}$ before optimizing problem (\ref{equ_MTFS_l21_objective}). Liu et al. \cite{lpz09} propose to use the $\ell_{\infty,1}$ norm to select features with the objective function formulated as
{\small
\begin{equation}
\min_{\mathbf{W},\mathbf{b}}\ \ L(\mathbf{W},\mathbf{b})+\lambda\|\mathbf{W}\|_{\infty,1}.
\label{equ_MTFS_l_infty_1_objective}
\end{equation}
}\noindent
A block coordinate descent method is proposed to solve problem (\ref{equ_MTFS_l_infty_1_objective}). In general, we can use the $\ell_{p,q}$ norm to select features for MTL.

In order to attain a more sparse subset of features, Gong et al. \cite{gyz13} propose a capped-$\ell_{p,1}$ regularizer for multi-task feature selection where $p=1$ or 2 and the objective function is formulated as
{\small
\begin{equation}
\min_{\mathbf{W},\mathbf{b}}\ \ L(\mathbf{W},\mathbf{b})+\lambda\sum_{i=1}^d\min(\|\mathbf{w}_i\|_p,\theta),
\label{equ_MSMTFL_objective}
\end{equation}
}\noindent
where $\mathbf{w}_i$ denotes the $i$th row of $\mathbf{W}$. With a given threshold $\theta$, the capped-$\ell_{p,1}$ regularizer (i.e., the second term in problem (\ref{equ_MSMTFL_objective})) focuses on rows with smaller $\ell_p$ norms than $\theta$, which is more likely to be sparse. When $\theta$ becomes large enough, the capped-$\ell_{p,1}$ regularizer becomes $\|\mathbf{W}\|_{p,1}$ and hence problem (\ref{equ_MSMTFL_objective}) degenerates to problem (\ref{equ_MTFS_l21_objective}) or (\ref{equ_MTFS_l_infty_1_objective}) when $p$ equals 2 or $\infty$.

Lozano and Swirszcz \cite{ls12} propose a multi-level Lasso for MTL where the $(j,i)$th entry in the parameter matrix $\mathbf{W}$ is defined as $w_{ji}=\theta_j\hat{w}_{ji}$. When $\theta_j$ is equal to 0, $w_{ji}$ becomes 0 for $i\in[m]$ and hence the $j$th feature is not selected by the model. In this sense, $\theta_j$ controls the global sparsity for the $j$th feature among the $m$ tasks. Moreover, when $\hat{w}_{ji}$ becomes 0, $w_{ji}$ is also 0 for $i$ only, implying that the $j$th feature is not useful for task $\mathcal{T}_i$, and so $\hat{w}_{ji}$ is a local indicator for the sparsity in task $\mathcal{T}_j$. Based on these observations, $\theta_j$ and $\hat{w}_{ji}$ are expected to be sparse, leading to the objective function formulated as
{\small
\begin{align}
\hskip -0.15in \min_{\bm{\theta},\mathbf{\hat{W}},\mathbf{b}} L(\mathbf{W},\mathbf{b})+\lambda_1\|\bm{\theta}\|_1+\lambda_2\|\mathbf{\hat{W}}\|_1\
\mathrm{s.t.}\ w_{ji}=\theta_j\hat{w}_{ji}, \theta_j\ge 0,
\label{equ_MLLasso_objective}
\end{align}
}\noindent
where $\bm{\theta}=(\theta_1,\ldots,\theta_d)^T$, $\mathbf{\hat{W}}=(\mathbf{\hat{w}}^1,\ldots,\mathbf{\hat{w}}^m)$, and the nonnegative constraint on $\theta_j$ is to keep the model identifiability. It has been proved in \cite{ls12} that problem (\ref{equ_MLLasso_objective}) leads to a regularizer $\sum_{j=1}^d\sqrt{\|\mathbf{w}_j\|_1}$, the square root of the $\ell_{1,\frac{1}{2}}$ norm regularization. Moreover, Wang et al. \cite{wbys14} extend problem (\ref{equ_MLLasso_objective}) to a general situation where the regularizer becomes $\lambda_1\sum_{i=1}^m\|\mathbf{\hat{w}}^i\|_p^p+\lambda_2\|\bm{\theta}\|_q^q$. By utilizing a priori information describing the task relations in a hierarchical structure, Han et al. \cite{hzsx14} propose a multi-component product based decomposition for $w_{ij}$ where the number of components in the decomposition can be arbitrary instead of only 2 in \cite{ls12,wbys14}. Similar to \cite{ls12}, Jebara \cite{jebara04} proposes to learn a binary indicator vector to do multi-task feature selection based on the maximum entropy discrimination formalism.

Similar to \cite{hzsx14} where a priori information is given to describe task relations in a hierarchical/tree structure, Kim and Xing \cite{kx10} utilize the given tree structure to design a regularizer on $\mathbf{W}$ as $f(\mathbf{W})=\sum_{i=1}^d\sum_{v\in V}\lambda_{v}\|\mathbf{w}_{i,G_v}\|_2$, where $V$ denotes the set of nodes in the given tree structure, $G_v$ denotes the set of leaf nodes (i.e., tasks) in a sub-tree rooted at node $v$, and $\mathbf{w}_{i,G_v}$ denotes a subvector of the $i$th row of $\mathbf{W}$ indexed by $G_v$. This regularizer not only enforces each row of $\mathbf{W}$ to be sparse as the $\ell_{2,1}$ norm did in problem (\ref{equ_MTFS_l21_objective}), but also induces the sparsity in subsets of each row in $\mathbf{W}$ based on the tree structure.

Different from conventional multi-task feature selection methods which assume that different tasks share a set of original features, Zhou et al. \cite{zjh10} consider a different scenario where useful features in different tasks have no overlapping. In order to achieve this, an exclusive Lasso model is proposed with the objective function formulated as
{\small
\begin{equation*}
\min_{\mathbf{W},\mathbf{b}}\ L(\mathbf{W},\mathbf{b})+\lambda\|\mathbf{W}\|_{1,2}^2,
\end{equation*}
}\noindent
where the regularizer is the squared $\ell_{1,2}$ norm on $\mathbf{W}$.

Another way to select common features for MTL is to use sparse priors to design probabilistic or Bayesian models. For $\ell_{p,1}$-regularized multi-task feature selection, Zhang et al. \cite{zyx10} propose a probabilistic interpretation where the $\ell_{p,1}$ regularizer corresponds to a generalized normal prior: $w_{ji}\sim\mathcal{GN}(\cdot|0,\rho_j,p)$, where $\cdot$ denotes a (random) variable when we do not want to introduce it explicitly. Based on this interpretation, Zhang et al. \cite{zyx10} further propose a probabilistic framework for multi-task feature selection, in which task relations and outlier tasks can be identified, based on the matrix-variate generalized normal prior.

In \cite{hh13}, a generalized horseshoe prior is proposed to do feature selection for MTL as:
{\small
\begin{equation*}
\mathbb{P}(\mathbf{w}^i)=\int\prod_{j=1}^d\mathcal{N}(w_{ji}|0,\frac{u_{ji}}{v_{ji}})\mathcal{N}(\mathbf{u}^i|0,\rho^2\mathbf{C})
\mathcal{N}(\mathbf{v}^i|0,\gamma^2\mathbf{C})\mathrm{d}\mathbf{u}^i\mathrm{d}\mathbf{v}^i,
\end{equation*}
}\noindent
where $\mathcal{N}(\cdot|\mathbf{m},\bm{\sigma})$ denotes a univariate or multivariate normal distribution with $\mathbf{m}$ as the mean and $\bm{\sigma}$ as the variance or covariance matrix, $u_{ji}$ and $v_{ji}$ are the $j$th entries in $\mathbf{u}^i$ and $\mathbf{v}^i$, respectively, and $\rho,\gamma$ are hyperparameters. Here $\mathbf{C}$ shared by all the tasks denotes the feature correlation matrix to be learned from data and it encodes an assumption that different tasks share identical feature correlations. When $\mathbf{C}$ becomes an identity matrix which means that features are independent, this prior degenerates to the horseshoe prior. %\cite{cps09}

Hern{\'{a}}ndez{-}Lobato et al. \cite{hhg15} propose a probabilistic model based on the horseshoe prior as
{\small
\begin{align}
\mathbb{P}(w_{ji})=&\left[\pi(w_{ji})^{\eta_{ji}}\delta_0^{1-\eta_{ji}}\right]^{z_j}
\left[\pi(w_{ji})^{\tau_{ji}}\delta_0^{1-\tau_{ji}}\right]^{\omega_i(1-z_j)}\nonumber\\
&\left[\pi(w_{ji})^{\gamma_j}\delta_0^{1-\gamma_j}\right]^{(1-\omega_i)(1-z_j)},\label{equ_DMTFL_prior}
\end{align}
}\noindent
where $\delta_0$ is the probability mass function at zero and $\pi(\cdot)$ denotes the density function of non-zero coefficients. In Eq. (\ref{equ_DMTFL_prior}), $z_j$ indicates whether feature $j$ is an outlier ($z_j=1$) or not
($z_j=0$) and $\omega_i$ indicates whether task $\mathcal{T}_i$ is an outlier ($\omega_i=1$) or not ($\omega_i=0$). Moreover, $\eta_{ji}$ and $\tau_{ji}$ indicate whether feature $j$ is relevant for the prediction in $\mathcal{T}_i$ ($\eta_{ji},\tau_{ji}=1$) or not ($\eta_{ji},\tau_{ji}=0$), and $\gamma_j$ indicates whether a non-outlier feature $j$ is relevant ($\gamma_j=1$) for the prediction or not ($\gamma_j=0$) in all non-outlier tasks. Based on the above definitions, the three terms in the right-hand side of Eq. (\ref{equ_DMTFL_prior}) specify probability density functions of $w_{ji}$ based on different situations of features and tasks. So this model can also handle outlier tasks but in a different way from \cite{zyx10}.

\subsubsection{Comparison between Two Sub-categories}

The two sub-categories have different characteristics where the feature transformation approach learns a transformation of the original features as the new representation but the feature selection approach selects a subset of the original features as the new representation for all the tasks. Based on the characteristics of those two approaches, the feature selection approach can be viewed as a special case of the feature transformation approach when the transformation matrix is a diagonal $0/1$ matrix where the diagonal entries with value 1 correspond to the selected features. By selecting a subset of the original features as the new representation, the feature selection approach has a better interpretability.

\subsection{Low-Rank Approach}

The relatedness among multiple tasks can imply the low-rank of $\mathbf{W}$, leading to the low-rank approach. For example, if the $i$th, $j$th and $k$th tasks are related in that the model parameter $\mathbf{w}^i$ of the $i$th task is a linear combination of those of the other two tasks, then it is easy to show that the rank of $\mathbf{W}$ is at most $m-1$ and hence of low rank. From this perspective, the more the relatedness is, the lower the rank of $\mathbf{W}$ is.

Ando and Zhang \cite{az05} assume that model parameters of different tasks share a low-rank subspace in part and specifically, $\mathbf{w}^i$ takes the following form as
{\small
\begin{equation}
\mathbf{w}^i=\mathbf{u}^i+\bm{\Theta}^T\mathbf{v}^i.\label{equ_ASO_parameter_decomposition}
\end{equation}
}\noindent
Here $\bm{\Theta}\in\mathbb{R}^{h\times d}$ is the shared low-rank subspace by multiple tasks where $h<d$. Then we can write in a matrix form as $\mathbf{W}=\mathbf{U}+\bm{\Theta}^T\mathbf{V}$. Based on the form of $\mathbf{W}$, the objective function proposed in \cite{az05} is formulated as
{\small
\begin{align}
\min_{\mathbf{U},\mathbf{V},\bm{\Theta},\mathbf{b}} L(\mathbf{U}+\bm{\Theta}^T\mathbf{V},\mathbf{b}) +\lambda\|\mathbf{U}\|_F^2\ \
\mathrm{s.t.}\ \bm{\Theta}\bm{\Theta}^T=\mathbf{I},\label{equ_ASO_objective}
\end{align}
}\noindent
where $\|\cdot\|_F$ denotes the Frobenius norm. The orthonormal constraint on $\bm{\Theta}$ in problem (\ref{equ_ASO_objective}) makes the subspace non-redundant. When $\lambda$ is large enough, the optimal $\mathbf{U}$ can become a zero matrix and hence problem (\ref{equ_ASO_objective}) is very similar to problem (\ref{equ_MTFL_objective}) except that there is no regularization on $\mathbf{V}$ in problem (\ref{equ_ASO_objective}) and that $\bm{\Theta}$ has a smaller number of rows than columns. Chen et al. \cite{ctly09} generalize problem (\ref{equ_ASO_objective}) as
{\small
\begin{eqnarray}
\hskip -0.27in\min_{\mathbf{U},\mathbf{V},\bm{\Theta},\mathbf{b}}&&L(\mathbf{W},\mathbf{b})
+\lambda_1\|\mathbf{U}\|_F^2+\lambda_2\|\mathbf{W}\|_F^2\nonumber\\
\mathrm{s.t.}&&\mathbf{W}=\mathbf{U}+\bm{\Theta}^T\mathbf{V},\ \bm{\Theta}\bm{\Theta}^T=\mathbf{I}.\label{equ_cASO_objective}
\end{eqnarray}
}\noindent
When setting $\lambda_2$ to be 0, problem (\ref{equ_cASO_objective}) reduces to problem (\ref{equ_ASO_objective}). Even though problem (\ref{equ_cASO_objective}) is non-convex, with some convex relaxation technique, it can be relaxed to the following convex problem as
{\small
\begin{align}
\hskip -0.15in \min_{\mathbf{W},\mathbf{b},\mathbf{M}}L(\mathbf{W},\mathbf{b})
+\lambda\mathrm{tr}\big(\mathbf{W}^T\left(\mathbf{M}+\eta\mathbf{I}\right)^{-1}\mathbf{W}\big)\
\mathrm{s.t.}\
{\mathrm{tr}(\mathbf{M})=h\atop
\mathbf{0}\preceq\mathbf{M}\preceq\mathbf{I}
},\label{equ_cASO_objective_relax}
\end{align}
}\noindent
where $\eta=\lambda_2/\lambda_1$ and $\lambda=\lambda_1\eta(\eta+1)$. One advantage of problem (\ref{equ_cASO_objective_relax}) over problem (\ref{equ_cASO_objective}) is that the global optimum of the convex problem (\ref{equ_cASO_objective_relax}) is much easier to be obtained than that of the non-convex problem (\ref{equ_cASO_objective}). Compared with the alternative objective function (\ref{equ_MTFL_objective_2}) in the MTFL method, problem (\ref{equ_cASO_objective_relax}) has a similar formulation where $\mathbf{M}$ models the feature covariance for all the tasks. Problem (\ref{equ_ASO_objective}) is extended in \cite{adg10} to a general case where different $\mathbf{w}^i$'s lie in a manifold instead of a subspace. Moreover, in \cite{zgy05}, a latent variable model is proposed for $\mathbf{W}$ with the same decomposition as Eq. (\ref{equ_ASO_parameter_decomposition}) and it can provide a framework for MTL by modeling more cases than problem (\ref{equ_ASO_objective}) such as task clustering, sharing sparse representation, duplicate tasks and evolving tasks.

It is well known that using the trace norm as a regularizer can make a matrix have low rank and hence this regularization is suitable for MTL. Specifically, an objective function with the trace norm regularization is proposed in \cite{ptjy10} as
{\small
\begin{equation}
\min_{\mathbf{W},\mathbf{b}}\ \ L(\mathbf{W},\mathbf{b})+\lambda\|\mathbf{W}\|_{S(1)},
\label{equ_TNR_objective}
\end{equation}
}\noindent
where $\mu_i(\mathbf{W})$ denotes the $i$th smallest singular value of $\mathbf{W}$ and $\|\mathbf{W}\|_{S(1)}=\sum_{i=1}^{\min(m,d)}\mu_i(\mathbf{W})$ denotes the trace norm of matrix $\mathbf{W}$. Based on the trace norm, Han and Zhang \cite{hz16} propose a capped trace regularizer with the objective function formulated as
{\small
\begin{equation}
\min_{\mathbf{W},\mathbf{b}}\ L(\mathbf{W},\mathbf{b})+\lambda\sum_{i=1}^{\min(m,d)}\min(\mu_i(\mathbf{W}),\theta).
\label{equ_CTNR_objective}
\end{equation}
}\noindent
With the use of the threshold $\theta$, the capped trace regularizer only penalizes small singular values of $\mathbf{W}$, which is related to the determination of the rank of $\mathbf{W}$. When $\theta$ is large enough, the capped trace regularizer will become the trace norm and hence problem (\ref{equ_CTNR_objective}) will reduce to problem (\ref{equ_TNR_objective}). Moreover, a spectral $k$-support norm is proposed in \cite{mps14} as an improvement over the trace norm regularization.

The trace norm regularization has been extended to regularize model parameters in deep multi-task models. Specifically, the weights in the last several fully connected layers of deep multi-task neural networks can be viewed as the parameters of learners for all the tasks. In this view, the weights connecting two consecutive layers for one task can be organized in a matrix and hence the weights of all the tasks can form a tensor. Based on such tensor representations, several tensor trace norms, which are based on the trace norm, are used in \cite{yh17b} as regularizers to identify the low-rank structure of the parameter tensor.

\subsection{Task Clustering Approach}

The task clustering approach assumes that different tasks form several clusters, each of which consists of similar tasks. As indicated by its name, this approach has a close connection with clustering algorithms and it can be viewed as an extension of clustering algorithms to the task level while the conventional clustering algorithms are on the data level.

Thrun and Sullivan \cite{to96} propose the first task clustering algorithm by using a weighted nearest neighbor classifier for each task, where the initial weights to define the weighted Euclidean distance are learned by minimizing pairwise within-class distances and maximizing pairwise between-class distances simultaneously within each task. Then a task transfer matrix $\mathbf{A}$ is defined with its $(i,j)$th entry $a_{ij}$ recording the generalization accuracy obtained for task $\mathcal{T}_i$ by using task $\mathcal{T}_j$'s distance metric. Based on $\mathbf{A}$, $m$ tasks can be grouped into $r$ clusters $\{\mathcal{C}_i\}_{i=1}^r$ by maximizing $\sum_{t=1}^r\frac{1}{|\mathcal{C}_t|}\sum_{i,j\in\mathcal{C}_t}a_{ij}$, where $|\cdot|$ denotes the cardinality of a set. After obtaining the cluster structure among all the tasks, the training data of tasks in a cluster will be pooled together to learn the final weighted nearest neighbor classifier. This approach has been extended to an iterative learning process \cite{cm12} in a similar way to $k$-means clustering.

Bakker and Heskes \cite{bh03} propose a multi-task Bayesian neural network model with the network structure similar to Fig. \ref{fig_MTLNN_sample} where input-to-hidden weights are shared by all the tasks but hidden-to-output weights are task-specific. By defining $\mathbf{w}^i$ as the vector of hidden-to-output weights for task $\mathcal{T}_i$, the multi-task Bayesian neural network assigns a mixture of Gaussian prior to it: $\mathbf{w}^i\sim\sum_{j=1}^r\pi_j\mathcal{N}(\cdot|\mathbf{m}_j,\bm{\Sigma}_j)$, where $\pi_j$, $\mathbf{m}_j$ and $\bm{\Sigma}_j$ specify the prior, the mean and the covariance in the $j$th cluster. For tasks in a cluster, they will share a Gaussian distribution. When $r$ equals 1, this model degenerates to a case where model parameters of different tasks share a prior, which is similar to several Bayesian MTL models such as \cite{yts05,yty07,lhrlc15} that are based on Gaussian processes and $t$ processes.

Xue et al. \cite{xlck07} deploy the Dirichlet process to do clustering on task level. Specifically, it defines the prior on $\mathbf{w}^i$ as
{\small
\begin{equation*}
\mathbf{w}^i\sim G,\ G\sim \mathcal{DP}(\alpha,G_0)\ \forall i\in[m],
\end{equation*}
}\noindent
where $\mathcal{DP}(\alpha,G_0)$ denotes a Dirichlet process with $\alpha$ as a positive scaling parameter and $G_0$ a base distribution. To see the clustering effect, by integrating out $G$, the conditional distribution of $\mathbf{w}^i$, given model parameters of other tasks $\mathbf{W}_{-i}=\{\cdots,\mathbf{w}^{i-1},\mathbf{w}^{i+1},\cdots\}$, is
{\small
\begin{equation*}
\mathbb{P}(\mathbf{w}^i|\mathbf{W}_{-i},\alpha,G_0)=\frac{\alpha}{m-1+\alpha}G_0+\frac{1}{m-1+\alpha}\sum_{j=1,j\ne i}^m \delta_{\mathbf{w}^j},
\end{equation*}
}\noindent
where $\delta_{\mathbf{w}^j}$ denotes the distribution concentrated at a single point $\mathbf{w}^j$. So $\mathbf{w}^i$ can be equal to either $\mathbf{w}^j$ ($j\ne i$) with probability $\frac{1}{m-1+\alpha}$, which corresponds to the case that those two tasks lie in the same cluster, or a new sample from $G_0$ with probability $\frac{\alpha}{m-1+\alpha}$, which is the case that task $\mathcal{T}_i$ forms a new task cluster. When $\alpha$ is large, the chance to form a new task cluster is large and so $\alpha$ affects the number of task clusters. This model is extended in \cite{xdc07,llc11} to a case where different tasks in a task cluster share useful features via a matrix stick-breaking process and a beta-Bernoulli hierarchical prior, respectively, and in \cite{qldc08} where each task is a compressive sensing task. Moreover, a nested Dirichlet process is proposed in \cite{ncd07,npcd08} to use Dirichlet processes to learn both task clusters and the state structure of an infinite hidden Markov model, which handles sequential data in each task. In \cite{prwd12}, $\mathbf{w}^i$ is decomposed as $\mathbf{w}^i=\mathbf{u}^i+\bm{\Theta}_i^T\mathbf{v}^i$ similar to Eq. (\ref{equ_ASO_parameter_decomposition}), where $\mathbf{u}^i$ and $\bm{\Theta}_i$ are sampled according to a Dirichlet process.

Different from \cite{bh03,xlck07}, Jacob et al. \cite{jbv08} aim to learn task clusters under the regularization framework by considering three orthogonal aspects, including a global penalty to measure on average how large the parameters, a measure of between-cluster variance to quantify the distance among different clusters and a measure of within-cluster variance to quantify the compactness of task clusters. By combining these three aspects and adopting some convex relaxation technique, a convex objective function is formulated as
{\small
\begin{align}
\min_{\mathbf{W},\mathbf{b},\bm{\Sigma}}\ & L(\mathbf{W},\mathbf{b})+\lambda\mathrm{tr}(\mathbf{W}\mathbf{1}\mathbf{1}^T\mathbf{W}^T)
+\mathrm{tr}(\mathbf{\tilde{W}}\bm{\Sigma}^{-1}\mathbf{\tilde{W}}^T)\nonumber\\
\mathrm{s.t.}\ &\mathbf{\tilde{W}}=\mathbf{W}\bm{\Pi},\ \alpha\mathbf{I}\preceq\bm{\Sigma}\preceq\beta\mathbf{I},\ \mathrm{tr}(\bm{\Sigma})=\gamma,\label{equ_CMTL_objective}
\end{align}
}\noindent
where $\bm{\Pi}$ denotes the $m\times m$ centering matrix, $\mathbf{1}$ denotes a column vector of all ones with its size depending on the context, and $\alpha,\beta,\gamma$ are hyperparameters.

Kang et al. \cite{kgs11} extend the MTFL method \cite{aep08} to the case with multiple task clusters and aim to minimize the squared trace norm in each cluster. A diagonal matrix, $\mathbf{Q}_i\in\mathbb{R}^{m\times m}$, is defined as a cluster indicator matrix for the $i$th cluster. The $j$th diagonal entry of $\mathbf{Q}_i$ is equal to 1 if task $\mathcal{T}_j$ lies in the $i$th cluster and otherwise 0. Since each task can belong to only one cluster, it is easy to see that $\sum_{i=1}^r\mathbf{Q}_i=\mathbf{I}$. Based on these considerations, the objective function is formulated as
{\small
\begin{align*}
\min_{\mathbf{W},\mathbf{b},\{\mathbf{Q}_i\}} L(\mathbf{W},\mathbf{b})+\lambda\sum_{i=1}^r\|\mathbf{W}\mathbf{Q}_i\|_{S(1)}^2\
\mathrm{s.t.}
{
\mathbf{Q}_i\in\{0,1\}^{m\times m}\atop
\sum_{i=1}^r\mathbf{Q}_i=\mathbf{I}
}.
\end{align*}
}\noindent
When $r$ equals 1, this method reduces to the MTFL method.

Han and Zhang \cite{hz15a} devise a structurally sparse regularizer to cluster tasks with the objective function as
{\small
\begin{equation}
\min_{\mathbf{W},\mathbf{b}}\ L(\mathbf{W},\mathbf{b})
+\lambda\sum_{j>i}\|\mathbf{w}^i-\mathbf{w}^j\|_2.\label{equ_MeTaG_single_level_objective}
\end{equation}
}\noindent
Problem (\ref{equ_MeTaG_single_level_objective}) is a special case of the method proposed in \cite{hz15a} with only one level of task clusters. The regularizer on $\mathbf{W}$ enforces any pair of columns in $\mathbf{W}$ to have a chance to be identical and after solving problem (\ref{equ_MeTaG_single_level_objective}), the cluster structure can be discovered by comparing columns in $\mathbf{W}$. One advantage of this structurally sparse regularizer is that the convex problem (\ref{equ_MeTaG_single_level_objective}) can automatically determine the number of task clusters.

Barzilai and Crammer \cite{bc15} propose a task clustering method by defining $\mathbf{W}$ as $\mathbf{W}=\mathbf{FG}$ where $\mathbf{F}\in\mathbb{R}^{d\times r}$ and $\mathbf{G}\in\{0,1\}^{r\times m}$. With an assumption that each task belongs to only one cluster, the objective function is formulated as
{\small
\begin{align}
\min_{\mathbf{F},\mathbf{G},\mathbf{b}} L(\mathbf{FG},\mathbf{b})
+\lambda\|\mathbf{F}\|_F^2\
\mathrm{s.t.}\ {\mathbf{G}\in\{0,1\}^{r\times m}\atop\|\mathbf{g}^i\|_2=1\ \forall i\in[m]},\label{equ_CMTC_objective}
\end{align}
}\noindent
where $\mathbf{g}^i$ denotes the $i$th column of $\mathbf{G}$. When using the hinge loss or logistic loss, this non-convex problem can be relaxed to a min-max problem, which has a global optimum, by utilizing the dual problem with respect to $\mathbf{W}$ and $\mathbf{b}$ and discarding some non-convex constraints.

Zhou and Zhao \cite{zz16} aim to cluster tasks by identifying representative tasks which are a subset of the given $m$ tasks. If task $\mathcal{T}_i$ is selected by task $\mathcal{T}_j$ as a representative task, then it is expected that model parameters for $\mathcal{T}_j$ are similar to those of $\mathcal{T}_i$. $z_{ij}$ is defined as the probability that task $\mathcal{T}_j$ selects task $\mathcal{T}_i$ as its representative task. Then based on a matrix $\mathbf{Z}$ whose $(i,j)$th entry is $z_{ij}$, the objective function is formulated as
{\small
\begin{align}
\hskip -0.05in \min_{\mathbf{W},\mathbf{b},\mathbf{Z}}& L(\mathbf{W},\mathbf{b})+\lambda_1\|\mathbf{W}\|_F^2
+\lambda_2\sum_{i=1}^m\sum_{j=1}^m z_{ij}\|\mathbf{w}^i-\mathbf{w}^j\|_2^2+\lambda_3\|\mathbf{Z}\|_{2,1}\nonumber\\
\mathrm{s.t.}\ &\mathbf{Z}\ge\mathbf{0},\ \mathbf{Z}^T\mathbf{1}=\mathbf{1}.\label{equ_FCMTL_objective}
\end{align}
}\noindent
The third term in the objective function of problem (\ref{equ_FCMTL_objective}) enforces the closeness of each pair of tasks based on $\mathbf{Z}$ and the last term employs the $\ell_{2,1}$ norm to enforce the row sparsity of $\mathbf{Z}$ which implies that the number of representative tasks is limited. The constraints in problem (\ref{equ_FCMTL_objective}) guarantee that entries in $\mathbf{Z}$ define valid probabilities. Problem (\ref{equ_FCMTL_objective}) is related to problem (\ref{equ_MeTaG_single_level_objective}) since the regularizer in problem (\ref{equ_MeTaG_single_level_objective}) can be reformulated as
$2\sum_{j>i}\|\mathbf{w}^i-\mathbf{w}^j\|_2=\min_{\mathbf{\hat{Z}}\ge\mathbf{0}}\sum_{j>i}\left(\hat{z}_{ij}\|\mathbf{w}^i-\mathbf{w}^j\|_2^2
+\frac{1}{\hat{z}_{ij}}\right)$, where both the regularizer and constraint on $\mathbf{\hat{Z}}$ are different from those on $\mathbf{Z}$ in problem (\ref{equ_FCMTL_objective}).

Previous studies assume that each task can belong to only one task cluster and this assumption seems too restrictive. In \cite{kd12}, a GO-MTL method relaxes this assumption by allowing a task to belong to more than one cluster and defines a decomposition of $\mathbf{W}$ similar to problem (\ref{equ_CMTC_objective}) as $\mathbf{W}=\mathbf{LS}$ where $\mathbf{L}\in\mathbb{R}^{d\times r}$ denotes the latent basis with $r<m$ and $\mathbf{S}\in\mathbb{R}^{r\times m}$ contains linear combination coefficients for all the tasks. $\mathbf{S}$ is assumed to be sparse since each task is generated from only a few columns in $\mathbf{L}$ or equivalently belongs to a small number of clusters. The objective function is formulated as
{\small
\begin{equation}
\min_{\mathbf{L},\mathbf{S},\mathbf{b}} L(\mathbf{LS},\mathbf{b})+\lambda_1\|\mathbf{S}\|_1+\lambda_2\|\mathbf{L}\|_F^2.
\label{equ_GO_MTL_objective}
\end{equation}
}\noindent
Compared with the objective function of multi-task sparse coding, i.e., problem (\ref{equ_MTSC_objective}), we can see that when the regularization parameters take appropriate values, these two problems are almost equivalent except that in multi-task sparse coding, the dictionary $\mathbf{U}$ is overcomplete, while here the number of columns in $\mathbf{S}$ is smaller than that of its rows. This method has been extended in \cite{yh17a} to decompose the parameter tensor in the fully connected layers of deep neural networks.

Among the aforementioned methods, the method in \cite{to96} first \mbox{identifies} the cluster structure and then learns the model parameters of all the tasks separately, which is not preferred since the cluster structure learned may be suboptimal for the model parameters, hence follow-up works learn model parameters and the cluster structure together. An important problem in clustering is to determine the number of clusters and this is also important for this approach. Out of the above methods, only methods in \cite{xlck07,hz15a} can automatically determine the number of task clusters, where the method in \cite{xlck07} depends on the capacity of the Dirichlet process while the method in \cite{hz15a} relies on the use of a structurally sparse regularizer. Among all those models, some belong to Bayesian learning, i.e., \cite{bh03,xlck07}, while the rest models are regularized models. Among those regularized methods, only the objective function proposed in \cite{hz15a} is convex while others are originally non-convex.

The task clustering approach is related to the low-rank approach. To see that, suppose that there are $r$ task clusters ($r<m$) and all the tasks in a cluster share the same model parameters, making the parameter matrix $\mathbf{W}$ low-rank with the rank at most $r$. From the perspective of modeling, by setting $\mathbf{u}^i$ to be a zero vector in Eq. (\ref{equ_ASO_parameter_decomposition}), we can see that the decomposition of $\mathbf{W}$ in \cite{az05} becomes similar to those in \cite{kd12,bc15}, which in some sense shows the relation between those two approaches. Moreover, the equivalence between problems (\ref{equ_cASO_objective_relax}) and (\ref{equ_CMTL_objective}), two typical methods in the low-rank and task clustering approaches, has been proved in \cite{zcy11}. The task clustering approach can visualize the learned cluster structure, which is an advantage over the low-rank approach.

\subsection{Task Relation Learning Approach}

In MTL, tasks are related and the task relatedness can be quantitated via task similarity, task correlation, task covariance and so on. Here we use task relations to include all the quantitative relatedness.

In earlier studies on MTL, task relations are assumed to be known as a priori information. In \cite{ep04,pw10}, each task is assumed to be similar to any other task and so model parameters of each task will be enforced to approach the average model parameters of all the tasks. In \cite{emp05,kksa07}, task similarities for each pair of tasks are given and these studies utilize the task similarities to design regularizers to guide the learning of multiple tasks in a principle that the more similar two tasks are, the closer the corresponding model parameters are expected to be. A similar formulation to \cite{emp05} is proposed in \cite{fgf14} to estimate the mean of multiple distributions by learning pairwise task relations and another similar formulation is proposed in \cite{yss16} for log-density gradient estimation. Given a tree structure describing relations among tasks in \cite{gwzksr11}, model parameters of a task corresponding to a node in the tree are enforced to be similar to those of its parent node.

However, in most applications, task relations are not available. In this case, learning task relations from data automatically is a good option. Bonilla et al. \cite{bcw07} propose a multi-task Gaussian process (MTGP) by defining a prior on $f^i_j$, the functional value for $\mathbf{x}^i_j$, as
$\mathbf{f}\sim\mathcal{N}(\cdot|\mathbf{0},\bm{\Sigma})$, where $\mathbf{f}=(f^1_1,\ldots,f^m_{n_m})^T$ contains the functional values for all the training data. $\bm{\Sigma}$, the covariance matrix, defines the covariance between $f^i_j$ and $f^p_q$ as
$\sigma(f^i_j,f^p_q)=\omega_{ip}k(\mathbf{x}^i_j,\mathbf{x}^p_q)$, where $k(\cdot,\cdot)$ denotes a kernel function and $\omega_{ip}$ describes the covariance between tasks $\mathcal{T}_i$ and $\mathcal{T}_p$. In order to keep $\bm{\Sigma}$ positive definite, a matrix $\bm{\Omega}$ containing $\omega_{ip}$ as its $(i,p)$th entry is also required to be positive definite, which makes $\bm{\Omega}$ the task covariance to describe the similarities between tasks. Then based on the Gaussian likelihood for labels given $\mathbf{f}$, the analytically marginal likelihood by integrating out $\mathbf{f}$ can be used to learn $\bm{\Omega}$ from data. In \cite{chai09}, the learning curve and generalization bound of the MTGP are studied. Since $\bm{\Omega}$ in MTGP has a point estimation which may lead to the overfitting, based on a proposed weight-space view of MTGP, Zhang and Yeung \cite{zy10a} propose a multi-task generalized $t$ process by placing an inverse-Wishart prior on $\bm{\Omega}$ as $\bm{\Omega}\sim\mathcal{IW}(\cdot|\nu,\bm{\Psi})$, where $\nu$ denotes the degree of freedom and $\bm{\Psi}$ is the base covariance for generating $\bm{\Omega}$. Since $\bm{\Psi}$ models the covariance between pairs of tasks, it can be determined based on the maximum mean discrepancy (MMD).

Different from \cite{bcw07,zy10a} which are Bayesian models, Zhang and Yeung \cite{zy10b,zy14} propose a regularized multi-task model called multi-task relationship learning (MTRL) by placing a matrix-variate normal prior on $\mathbf{W}$ as
{\small
\begin{equation}
\mathbf{W}\sim\mathcal{MN}(\cdot|\mathbf{0},\mathbf{I},\bm{\Omega}),\label{equ_MTRL_prior}
\end{equation}
}\noindent
where $\mathcal{MN}(\cdot|\mathbf{M},\mathbf{A},\mathbf{B})$ denotes a matrix-variate normal distribution with $\mathbf{M}$ as the mean, $\mathbf{A}$ the row covariance, and $\mathbf{B}$ the column covariance. Based on this prior as well as some likelihood function, the objective function for a modified maximum a posterior solution is formulated as
{\small
\begin{align}
\min_{\mathbf{W},\mathbf{b},\bm{\Omega}}\ & L(\mathbf{W},\mathbf{b})+\lambda_1\|\mathbf{W}\|_F^2
+\lambda_2\mathrm{tr}(\mathbf{W}\bm{\Omega}^{-1}\mathbf{W}^T)\nonumber\\
\mathrm{s.t.}\ &\bm{\Omega}\succ\mathbf{0},\ \mathrm{tr}(\bm{\Omega})\le 1,\label{equ_MTRL_objective}
\end{align}
}\noindent
where the second term in the objective function is to penalize the complexity of $\mathbf{W}$, the last term is due to the matrix-variate normal prior, and the constraints control the complexity of the positive definite covariance matrix $\bm{\Omega}$. It has been proved in \cite{zy10b,zy14} that problem (\ref{equ_MTRL_objective}) is jointly convex with respect to $\mathbf{W}$, $\mathbf{b}$ and $\bm{\Omega}$. Problem (\ref{equ_MTRL_objective}) has been extended to multi-task boosting \cite{zy12b} and multi-label learning \cite{zy13b} by learning label correlations. Problem (\ref{equ_MTRL_objective}) can also been interpreted from the perspective of reproducing kernel Hilbert spaces for vector-valued functions \cite{dogp11,cmpr15,crv15,jlhs15}. Moreover, Problem (\ref{equ_MTRL_objective}) is extended to learn sparse task relations in \cite{zy17} via the $\ell_1$ regularization on $\bm{\Omega}$ when the number of tasks is large. A model similar to problem (\ref{equ_MTRL_objective}) is proposed in \cite{zs10} via a matrix-variate normal prior on $\mathbf{W}$: $\mathbf{W}\sim\mathcal{MN}(\cdot|\mathbf{0},\bm{\Omega}_1,\bm{\Omega}_2)$, where $\bm{\Omega}_1^{-1}$ and $\bm{\Omega}_2^{-1}$ are assumed to be sparse. The MTRL model is extended in \cite{agz11} to use the symmetric matrix-variate generalized hyperbolic distribution to learn block sparse structure in $\mathbf{W}$ and in \cite{ylz13} to use the matrix generalized inverse Gaussian prior to learn low-rank $\bm{\Omega}_1$ and $\bm{\Omega}_2$. Moreover, the MTRL model is generalized to the multi-task feature selection problem \cite{zyx10} by learning task relations via the matrix-variate generalized normal distribution. Since the prior defined in Eq. (\ref{equ_MTRL_prior}) implies that $\mathbf{W}^T\mathbf{W}$ follows a Wishart distribution as $\mathcal{W}(\cdot|\mathbf{0},\bm{\Omega})$, Zhang and Yeung \cite{zy13a} generalize it as
{\small
\begin{equation}
(\mathbf{W}^T\mathbf{W})^t\sim\mathcal{W}(\cdot|\mathbf{0},\bm{\Omega}),\label{equ_MTHOL_prior}
\end{equation}
}\noindent
where $t$ is a positive integer to model high-order task relationships. Eq. (\ref{equ_MTHOL_prior}) can induce a new prior, which is a generalization of the matrix-variate normal distribution, on $\mathbf{W}$ and based on this new prior, a regularized method is devised to learn high-order task relations in \cite{zy13a}. The MTRL model has been extended to multi-output regression \cite{rkd12,ylz13,rlbs13,gzb16} by modeling the structure contained in noises via some matrix-variate priors. For deep neural networks, the MTRL method has been extended in \cite{lcwy17} by placing a tensor-variate normal distribution as a prior on the parameter tensor in the fully connected layers.

Different from the aforementioned methods which investigate the use of global learning models in MTL, Zhang \cite{zhang13} aims to learn the task relations in local learning methods such as the $k$-nearest-neighbor ($k$NN) classifier by defining the learning function as a weighted voting of neighbors:
{\small
\begin{equation}
f(\mathbf{x}^i_j)=\sum_{(p,q)\in N_k(i,j)}\sigma_{ip}s(\mathbf{x}^i_j,\mathbf{x}^p_q)y^p_q,\label{equ_MTKNN_learn_function}
\end{equation}
}\noindent
where $N_k(i,j)$ denotes the set of task indices and instance indices for the $k$ nearest neighbors of $\mathbf{x}^i_j$, i.e., $(p,q)\in N_k(i,j)$ meaning that $\mathbf{x}^p_q$ is one of the $k$ nearest neighbors of $\mathbf{x}^i_j$, $s(\mathbf{x}^i_j,\mathbf{x}^p_q)$ defines the similarity between $\mathbf{x}^i_j$ and $\mathbf{x}^p_q$, and $\sigma_{ip}$ represents the contribution of task $\mathcal{T}_p$ to $\mathcal{T}_i$ when $\mathcal{T}_p$ has some data points to be neighbors of a data point in $\mathcal{T}_i$. $\sigma_{ip}$ can be viewed as the similarity from $\mathcal{T}_p$ to $\mathcal{T}_i$. When $\sigma_{ip}=1$ for all $i$ and $p$, Eq. (\ref{equ_MTKNN_learn_function}) reduces to the decision function of the $k$NN classifier for all the tasks. Then the objective function to learn $\bm{\Sigma}$, which is a $m\times m$ matrix with $\sigma_{ip}$ as its $(i,p)$th entry, can be formulated as
{\small
\begin{eqnarray}
\min_{\bm{\Sigma}} && \sum_{i=1}^m\frac{1}{n_i}\sum_{j=1}^{n_i}l(y^i_j,f(\mathbf{x}^i_j))+\frac{\lambda_1}{4}\|\bm{\Sigma}-\bm{\Sigma}^T\|_F^2+\frac{\lambda_2}{2}\|\bm{\Sigma}\|_F^2\nonumber\\
\mathrm{s.t.} && \sigma_{ii}\ge 0\ \forall i\in[m], -\sigma_{ii}\le\sigma_{ij}\le\sigma_{ii}\ \forall i\ne j.\label{equ_MTKNN_objective}
\end{eqnarray}
}\noindent
The first regularizer in problem (\ref{equ_MTKNN_objective}) enforces $\bm{\Sigma}$ to be nearly symmetric and the second one is to penalize the complexity of $\bm{\Sigma}$. The constraints in problem (\ref{equ_MTKNN_objective}) guarantee that the similarity from one task to itself is positive and also the largest. Similarly, a multi-task kernel regression is proposed in \cite{zhang13} for regression tasks.

While the aforementioned methods whose task relations are symmetric except \cite{zhang13}, Lee et al. \cite{lyh16} focus on learning asymmetric task relations. Since different tasks are assumed to be related, $\mathbf{w}_i$ can lie in the space spanned by $\mathbf{W}$, i.e., $\mathbf{w}_i\approx\mathbf{W}\mathbf{a}_i$, and hence we have $\mathbf{W}\approx\mathbf{W}\mathbf{A}$. Here matrix $\mathbf{A}$ can be viewed as asymmetric task relations between pairs of tasks. By assuming that $\mathbf{A}$ is sparse, the objective function is formulated as
{\small
\begin{align}
\hskip -0.1in\min_{\mathbf{W},\mathbf{b},\mathbf{A}}\ & \sum_{i=1}^m(1+\lambda_1\|\mathbf{\hat{a}}_i\|_1)\sum_{j=1}^{n_i}l(y^i_j,(\mathbf{w}^i)^T\mathbf{x}^i_j+b_i)
+\lambda_2\|\mathbf{W}-\mathbf{W}\mathbf{A}\|_F^2\nonumber\\
\mathrm{s.t.}\ & a_{ij}\ge 0\ \forall i,j\in [m],\label{equ_AMTRL_objective}
\end{align}
}\noindent
where $\mathbf{\hat{a}}_i$ denotes the $i$th row of $\mathbf{A}$ by deleting $a_{ii}$. The term before the training loss of each task, i.e., $1+\lambda_1\|\mathbf{\hat{a}}_i\|_1$, not only enforces $\mathbf{A}$ to be sparse but also allows asymmetric information sharing from easier tasks to difficult ones. The regularizer in problem (\ref{equ_AMTRL_objective}) can make $\mathbf{W}$ approach $\mathbf{W}\mathbf{A}$ with the closeness depending on $\lambda_2$. To see the connection between problems (\ref{equ_AMTRL_objective}) and (\ref{equ_MTRL_objective}), we rewrite the regularizer in problem (\ref{equ_AMTRL_objective}) as
$\|\mathbf{W}-\mathbf{W}\mathbf{A}\|_F^2=\mathrm{tr}\left(\mathbf{W}(\mathbf{I}-\mathbf{A})(\mathbf{I}-\mathbf{A})^T\mathbf{W}^T\right)$.
Based on this reformulation, the regularizer in problem (\ref{equ_AMTRL_objective}) is a special case of that in problem (\ref{equ_MTRL_objective}) by assuming $\bm{\Omega}^{-1}=(\mathbf{I}-\mathbf{A})(\mathbf{I}-\mathbf{A})^T$. Though $\mathbf{A}$ is asymmetric, from the perspective of the regularizer, the task relations here are symmetric and act as the task precision matrix with a restrictive form.

\subsection{Decomposition Approach}

The decomposition approach assumes that the parameter matrix $\mathbf{W}$ can be decomposed into two or more component matrices $\{\mathbf{W}_k\}_{k=1}^h$ where $h\ge 2$, i.e., $\mathbf{W}=\sum_{k=1}^h\mathbf{W}_k$. The objective functions of most methods in this approach can be unified as
{\small
\begin{equation}
\min_{\{\mathbf{W}_i\}\in\mathcal{C}_W,\mathbf{b}}\ L\Big(\sum_{k=1}^h\mathbf{W}_k,\mathbf{b}\Big)
+\sum_{k=1}^h g_k(\mathbf{W}_k),\label{equ_DM_objective}
\end{equation}
}\noindent
where the regularizer is decomposable with respect to $\mathbf{W}_k$'s and $\mathcal{C}_W$ denotes a set of constraints for component matrices. To help understand problem (\ref{equ_DM_objective}), we introduce several instantiations as follows.

In \cite{jrsr10} where $h$ equals 2 and $\mathcal{C}_W=\emptyset$ is an empty set, $g_1(\cdot)$ and $g_2(\cdot)$ are defined as
{\small
\begin{equation*}
g_1(\mathbf{W}_1)=\lambda_1\|\mathbf{W}_1\|_{\infty,1},\ g_2(\mathbf{W}_2)=\lambda_2\|\mathbf{W}_2\|_1,
\end{equation*}
}\noindent
where $\lambda_1$ and $\lambda_2$ are positive regularization parameters. Similar to problem (\ref{equ_MTFS_l_infty_1_objective}), each row of $\mathbf{W}_1$ is likely to be a zero row and hence $g_1(\mathbf{W}_1)$ can help select important features. Due to the $\ell_1$ norm regularization, $g_2(\mathbf{W}_2)$ makes $\mathbf{W}_2$ sparse. Because of the characteristics of two regularizers, the parameter matrix $\mathbf{W}$ can eliminate unimportant features for all the tasks when the corresponding rows in both $\mathbf{W}_1$ and $\mathbf{W}_2$ are sparse. Moreover, $\mathbf{W}_2$ can identify features for tasks which have their own useful features that may be outliers for other tasks. Hence this model can be viewed as a `robust' version of problem (\ref{equ_MTFS_l_infty_1_objective}).

With two component matrices, Chen et al. \cite{cly10} define
{\small
\begin{align}
g_2(\mathbf{W}_2)=\lambda_2\|\mathbf{W}_2\|_1,\ \mathcal{C}_W=\{\mathbf{W}_1|\|\mathbf{W}_1\|_{S(1)}\le\lambda_1\},\label{equ_RTRN1_U_V_formulation}
\end{align}
}\noindent
where $g_1(\mathbf{W}_1)=0$. Similar to problem (\ref{equ_TNR_objective}), $\mathcal{C}_W$ makes $\mathbf{W}_1$ low-rank. With a sparse regularizer $g_2(\mathbf{W}_2)$, $\mathbf{W}_2$ makes the entire model matrix $\mathbf{W}$ more robust to outlier tasks in a way similar to the previous model. When $\lambda_2$ is large enough, $\mathbf{W}_2$ will become a zero matrix and then problem (\ref{equ_RTRN1_U_V_formulation}) will act similarly to problem (\ref{equ_TNR_objective}).

$g_i(\cdot)$'s in \cite{czy11} where $\mathcal{C}_W=\emptyset$ are defined as
{\small
\begin{equation}
g_1(\mathbf{W}_1)=\lambda_1\|\mathbf{W}_1\|_{S(1)},\quad g_2(\mathbf{W}_2)=\lambda_2\|\mathbf{W}_2^T\|_{2,1}.\label{equ_RTRN2_U_V_formulation}
\end{equation}
}\noindent
Different from the above two models which assume that $\mathbf{W}_2$ is sparse, here $g_2(\mathbf{W}_2)$ enforces $\mathbf{W}_2$ to be column-sparse. For related tasks, their columns in $\mathbf{W}_1$ are correlated via the trace norm regularization and the corresponding columns in $\mathbf{W}_2$ are zero. For outlier tasks which are unrelated to other tasks, the corresponding columns in $\mathbf{W}_2$ can take arbitrary values and hence model parameters in $\mathbf{W}$ for them have no low-rank structure even though those in $\mathbf{W}_1$ may have.

In \cite{gyz12}, these functions are defined as
{\small
\begin{equation}
g_1(\mathbf{W}_1)=\lambda_1\|\mathbf{W}_1\|_{2,1},\  g_2(\mathbf{W}_2)=\lambda_2\|\mathbf{W}_2^T\|_{2,1},\ \mathcal{C}_W=\emptyset.\label{equ_RMTFS_U_V_formulation}
\end{equation}
}\noindent
Similar to problem (\ref{equ_MTFS_l21_objective}), $g_1(\mathbf{W}_1)$ makes $\mathbf{W}_1$ row-sparse. Here $g_2(\mathbf{W}_2)$ is identical to that in \cite{czy11} and it makes $\mathbf{W}_2$ column-sparse. Hence $\mathbf{W}_1$ helps select useful features while non-zero columns in $\mathbf{W}_2$ capture outlier tasks.

With $h=2$, Zhong and Kwok \cite{zk12} define
{\small
\begin{equation*}
g_1(\mathbf{W}_1)=\lambda_1c(\mathbf{W}_1)+\lambda_2\|\mathbf{W}_1\|_F^2,g_2(\mathbf{W}_2)=\lambda_3\|\mathbf{W}_2\|_{F}^2,
\mathcal{C}_W=\emptyset,
\end{equation*}
}\noindent
where $c(\mathbf{U})=\sum_{i=1}^d\sum_{k>j}|u_{ij}-u_{ik}|$ with $u_{ij}$ as the $(i,j)$th entry in a matrix $\mathbf{U}$. Due to the sparse nature of the $\ell_1$ norm, $c(\mathbf{W}_1)$ enforces corresponding entries in different columns of $\mathbf{W}_1$ to be identical, which is equivalent to clustering tasks in terms of individual model parameters. Both the squared Frobenius norm regularizations in $g_1(\mathbf{W}_1)$ and $g_2(\mathbf{W}_2)$ penalize the complexities of $\mathbf{W}_1$ and $\mathbf{W}_2$. The use of $\mathbf{W}_2$ improves the model flexibility when not all the tasks exhibit a clear cluster structure.

Different from the aforementioned methods which have only two component matrices, an arbitrary number of component matrices are considered in \cite{zw13} with
{\small
\begin{align}
g_k(\mathbf{W}_k)=\lambda\big[(h-k)\|\mathbf{W}_k\|_{2,1}+(k-1)\|\mathbf{W}_k\|_1\big]/(h-1),\label{equ_MTC_objective}
\end{align}
}\noindent
where $\mathcal{C}_W=\emptyset$. According to Eq. (\ref{equ_MTC_objective}), $\mathbf{W}_k$ is assumed to be both sparse and row-sparse for all $k\in [h]$. Based on different regularization parameters on the regularizer of $\mathbf{W}_k$, we can see that when $k$ increases, $\mathbf{W}_k$ is more likely to be sparse than to be row-sparse. Even though each $\mathbf{W}_k$ is sparse or row-sparse, the entire parameter matrix $\mathbf{W}$ can be non-sparse and hence this model can discover the latent sparse structure among tasks.

In the above methods, different component matrices have no direct connection. When there is a dependency among component matrices, problem (\ref{equ_DM_objective}) can model more complex structure among tasks. For example, Han and Zhang \cite{hz15b} define
{\small
\begin{align}
&g_k(\mathbf{W}_k)=\lambda\sum_{i>j}\|\mathbf{w}_k^i-\mathbf{w}_k^j\|_2/\eta^{k-1}\ \forall k\in[h]\nonumber\\%\label{equ_TAT_objective}
&\mathcal{C}_W=\{\{\mathbf{W}_k\}|\ |\mathbf{w}_{k-1}^i-\mathbf{w}_{k-1}^j|\ge |\mathbf{w}_k^i-\mathbf{w}_k^j|\ \forall k\ge 2,\ \forall i>j\},\nonumber
\end{align}
}\noindent
where $\mathbf{w}_k^i$ denotes the $i$th column of $\mathbf{W}_k$. Note that the constraint set $\mathcal{C}_W$ relates component matrices and the regularizer $g_k(\mathbf{W}_k)$ makes each pair of $\mathbf{w}_k^i$ and $\mathbf{w}_k^j$ have a chance to become identical. Once this happens for some $i$, $j$, $k$, then based on the constraint set $\mathcal{C}_W$, $\mathbf{w}_{k'}^i$ and $\mathbf{w}_{k'}^j$ will always have the same value for $k'\ge k$. This corresponds to sharing all the ancestor nodes for two internal nodes in a tree and hence this method can learn a hierarchical structure to characterize task relations. When the constraints are removed, this method reduces to the multi-level task clustering method \cite{hz15a}, which is a generalization of problem (\ref{equ_MeTaG_single_level_objective}).

Another way to relate different component matrices is to use a non-decomposable regularizer as \cite{jn12} did, which is slightly different from problem (\ref{equ_DM_objective}) in terms of the regularizer. Specifically, given $m$ tasks, there are $2^m-1$ possible and non-empty task clusters. All the task clusters can be organized in a tree, where the root node represents a dummy node, nodes in the second level represent groups with a single task, and the parent-child relations are the `subset of' relation. In total, there are $h\equiv2^m$ component matrices each of which corresponds to a node in the tree and hence an index $a$ is used to denote both a level and the corresponding node in the tree. The objective function is formulated as
{\small
\begin{align}
\min_{\{\mathbf{W}_i\},\mathbf{b}}\ &L\left(\sum_{k=1}^h\mathbf{W}_k,\mathbf{b}\right)
+\left(\sum_{v\in V}\lambda_v\left(\sum_{a\in D(v)}r(\mathbf{W}_a)^{p}\right)^{\frac{1}{p}}\right)^2\nonumber\\
\mathrm{s.t.}\ & \mathbf{w}_a^i=\mathbf{0}\ \forall i\notin t(a),\label{equ_MLRMTL_objective}
\end{align}
}\noindent
where $p$ takes a value between 1 and 2, $D(a)$ denotes the set of all the descendants of $a$, $t(a)$ denotes the set of tasks contained in node $a$, $\mathbf{w}_a^i$ denotes the $i$th column of $\mathbf{W}_a$, and $r(\mathbf{W}_a)$ reflects relations among tasks in node $a$ based on $\mathbf{W}_a$. The regularizer in problem (\ref{equ_MLRMTL_objective}) is used to prune the subtree rooted at each node $v$ based on the $\ell_p$ norm. The constraint in problem (\ref{equ_MLRMTL_objective}) implies that for tasks not contained in a node $a$, the corresponding columns in $\mathbf{W}_a$ are zero. In \cite{jn12}, $r(\mathbf{W}_a)$ adopts the regularizer proposed in \cite{ep04} which enforces the parameters of all the tasks to approach their average.

Different from deep MTL models which are deep in terms of layers of feature representations, the decomposition approach can be viewed as a `deep' approach in terms of model parameters while most of previous approaches are just shallow ones, making this approach have more powerful capacity. Moreover, the decomposition approach can reduce to other approaches such as the feature learning, low-rank and task clustering approaches when there is only one component matrix and hence it can be considered as an improved version of those approaches.

\begin{table*}[!htb]
\caption{The performance comparison of representative MTL models in the five approaches on benchmark datasets in terms of some evaluation metric. nMSE stands for `normalized mean squared error', RMSE is for `root mean squared error', and AUC stands for `Area Under Curve'. $\uparrow$ after the evaluation metric implies that the larger value the better performance and $\downarrow$ indicates the opposite case.} \label{table-datasets}\centering
\resizebox{0.85\textwidth}{!}{
\begin{tabular}{|c|c|c|c|c|c|c|c|}
\hline
\multirow{2}{*}{Dataset (Reference)} & \multirow{2}{*}{Evaluation Metric} & \multirow{2}{*}{STL} &Feature Learning & Low-Rank & Task Clustering & Task Relation Learning & Decomposition \\
\cline{4-8}
& & & \cite{aep08} & \cite{ptjy10} & \cite{jbv08}/\cite{kgs11}/\cite{kd12}/\cite{yh17a} & \cite{zy10b}/\cite{jlhs15}/\cite{lcwy17} & \cite{jrsr10}/\cite{cly10}/\cite{czy11}/\cite{jn12}/\cite{hz15b} \\
\hline
School (\cite{hz15b}) & nMSE$\downarrow$ & --- & 0.4393 & --- & 0.4374/-/0.6466/- & --- & 0.4445/-/-/-/0.4169 \\
\hline
SARCOS  (\cite{czy11}) & nMSE$\downarrow$ & 0.1821 & 0.1568 & 0.1531 & --- & --- & 0.1495/0.1456/-/-/- \\ \hline
Computer Survey (\cite{zk12}) & RMSE$\downarrow$ & 2.381 & --- & --- & 2.072/-/-/- & 2.110/-/- & 2.138/2.052/2.074/-/- \\
\hline
Parkinson (\cite{jlhs15}) & Explained Variance$\uparrow$ & 2.8\% & --- & --- & 2.7\%/33.6\%/-/- & 12.0\%/27.0\%/- & -/-/-/16.8\%/- \\
\hline
Sentiment (\cite{zy10b}) & Classification Error$\downarrow$ & 0.2779 & 0.2756 & --- & --- & 0.2324/-/- & --- \\
\hline
MHC-I (\cite{zz16}) & Classification Error$\downarrow$ & 0.2010 & --- & --- & 0.1890/0.2050/-/- & 0.1870/-/- & 0.2030/-/0.2070/-/- \\
\hline
Landmine (\cite{jlhs15}) & AUC$\uparrow$ & 74.6\% & --- & --- & 75.9\%/76.7\%/-/- & 76.1\%/76.8\%/- & -/-/-/76.4\%/- \\
\hline
Office-Caltech (\cite{lcwy17}) & Classification Error$\downarrow$ & 0.0920 & 0.0740 & --- & -/-/-/0.0670 & 0.0690/-/0.0450 & -/-/0.0760/-/- \\
\hline
Office-Home (\cite{lcwy17})  & Classification Error$\downarrow$ & 0.3430 & 0.4170 & --- & -/-/-/0.3350 & 0.4070/-/0.3310 & -/-/0.4140/-/- \\
\hline
ImageCLEF (\cite{lcwy17}) & Classification Error$\downarrow$ & 0.3640 & 0.3440 & --- & -/-/-/0.2780 & 0.3350/-/0.2470 & -/-/0.3510/-/- \\
\hline
\end{tabular}
}
\end{table*}

\subsection{Comparisons among Different Approaches}

Based on the above introduction, we can see that different approaches exhibit their own characteristics. Specifically, the feature learning approach can learn common features, which are generic and invariant to all the tasks at hand and even new tasks, for all the tasks. When there exist outlier tasks which are unrelated to other tasks, the learned features can be influenced by outlier tasks significantly and they may cause the performance deterioration. By assuming that the parameter matrix is low-rank, the low-rank approach can explicitly learn the subspace of the parameter matrix or implicitly achieve that via some convex or non-convex regularizer. This approach is powerful but it seems applicable to only linear models, making nonlinear extensions non-trivial to be devised. The task clustering approach performs clustering on the task level in terms of model parameters and it can identify task clusters each of which consists of similar tasks. A major limitation of the task clustering approach is that it can capture positive correlations among tasks in the same cluster but ignore negative correlations among tasks in different clusters. Moreover, even though some methods in this category can automatically determine the number of clusters, most of them still need a model selection method such as cross validation to determine it, which may bring additional computational costs. The task relation learning approach can learn model parameters and pairwise task relations simultaneously. The learned task relations can give us insights about the relations between tasks and hence they improve the interpretability. The decomposition approach can be viewed as extensions of other parameter-based approaches by equipping multi-level parameters and hence they can model more complex task structure, e.g., tree structure. The number of components in the decomposition approach is important to the performance and needs to be carefully determined.

\subsection{Benchmark Datasets and Performance Comparison}
\label{sec:dataset}

In this section, we introduce some benchmark datasets for MTL and compare the performance of different MTL models on them.

Some benchmark datasets for MTL are listed as follows.
\begin{itemize}

\item School dataset \cite{bh03}: This dataset is to estimate examination scores of 15,362 students from 139 secondary schools in London from 1985 to 1987 where each school is treated as a task.  The input consists of four school-specific and three student-specific attributes.

\item SARCOS dataset\footnote{\url{http://www.gaussianprocess.org/gpml/data/}}: This dataset studies a multi-output problem of learning the inverse dynamics of 7 SARCOS anthropomorphic robot arms, each of which corresponds to a task, based on 21 features, including seven joint positions, seven joint velocities and seven joint accelerations. This dataset contains 48,933 data points.

\item Computer Survey dataset \cite{ampy07}: This dataset is taken from a survey of 180 persons/tasks who rated the likelihood of purchasing one of 20 different personal computers, resulting in 36,000 data points in all the tasks. The features contain 13 different computer characteristics (e.g., price, CPU and RAM) while the output is an integer rating on the scale 0-10.

\item Parkinson dataset \cite{jn12}: This dataset is to predict the disease symptom score of Parkinson for patients at different times using 19 bio-medical features. This dataset has 5,875 data points for 42 patients, each of whom is treated as a task.

\item Sentiment dataset\footnote{\url{http://www.cs.jhu.edu/~mdredze/datasets/sentiment/}}: This dataset is to classify reviews of four products/tasks, i.e., books, DVDs, electronics and kitchen appliances, from Amazon into two classes: positive and negative reviews. For each task, there are 1,000 positive and 1,000 negative reviews, respectively. %Each data point has 473856 feature dimensions.

\item MHC-I dataset \cite{jbv08}: This databset contains binding affinities of 15,236 peptides with 35 MHC-I molecules. Each MHC-I molecule is considered as a task and the goal is to predict whether a peptide binds a molecule. %resulting in 15236 points involving 35 different molecules

\item Landmine dataset \cite{xlck07}: This dataset consists of 9-dimensional data points, whose features are extracted from radar images, from 29 landmine fields/tasks. Each task is to classify a data point into two classes (landmine or clutter). There are 14,820 data points in total.

\item Office-Caltech dataset \cite{gssg12}: The dataset contains data from 10 common categories shared in the Caltech-256 dataset and the Office dataset which consists of images collected from three distinct domains/tasks: Amazon, Webcam and DSLR, making this dataset contain 4 tasks. There are 2,533 images in all the tasks.

\item Office-Home dataset\footnote{\url{http://hemanthdv.org/OfficeHome-Dataset}}: This dataset consists of images from 4 different domains/tasks: artistic images, clip art, product images and real-world images. Each task contains images of 65 object categories collected in the office and home settings. In total, there are about 15,500 images in all the tasks.

\item ImageCLEF dataset\footnote{\url{http://imageclef.org/2014/adaptation}}: This dataset contains 12 common categories shared by four tasks: Caltech-256, ImageNet ILSVRC 2012, Pascal VOC 2012 and Bing. There are about 2,400 images in all the tasks. %All three datasets are evaluated using DeCAF7 features for shallow methods and original images for deep methods.

\end{itemize}

In the above benchmark datasets, the first four datasets consist of regression tasks while the other datasets are classification tasks, where each task in the Sentiment, MHC-I and Landmine datasets is a binary classification problem and that in the other three image datasets is a multi-class classification problem. In order to compare different MTL approaches on those benchmark datasets, we select some representative MTL methods from each of the five approaches introduced in the previous sections and list in Table \ref{table-datasets} their performance reported in the MTL literature. We also include the performance of Single-Task Learning (STL), which trains a learning model for each task separately, for comparison. It is easy to see that MTL models perform better than STL counterparts in most cases, which verifies the effectiveness of MTL. Usually, different datasets have their own characteristics, making them more suitable for some MTL approach. For example, according to the studies in \cite{bh03,emp05,zk12}, different tasks in the School dataset are found to be very similar to each other. According to \cite{xlck07}, the Landmine dataset can have two task clusters, where the first cluster consisting of the first 15 tasks corresponds to regions that are relatively highly foliated and the rest tasks belong to another cluster with regions that are bare earth or deserts. According to \cite{jbv08}, it is well known in the vaccine design community that some molecules/tasks in the MHC-I dataset can be grouped into empirically defined supertypes known to have similar binding behaviors. For those three datasets, according to Table \ref{table-datasets} we can see that the task clustering, task relation learning and decomposition approaches have better performance since they can identify the cluster structure contained in the data in a plain or hierarchical way. For other datasets, they do not have so obvious structure among tasks but some MTL models can learn task correlations, which can bring more insights for model design and the interpretation of experimental results. For example, the task correlations in the SARCOS and Sentiment datasets are shown in Tables 2 and 3 of \cite{zy10b}, and the task similarities in the Office-Caltech dataset are shown in Figure 3(b) of \cite{lcwy17}. Moreover, for image datasets (i.e., Office-Caltech, Office-Home and ImageCLEF), deep MTL models (e.g., \cite{yh17a,lcwy17}) achieve better performance than shallow models since they can learn powerful feature representations, while the rest datasets are from diverse areas, making shallow models perform well on them.

\subsection{Another Taxonomy for Regularized MTL Methods}
\label{sec:another_taxonomy}

Regularized methods form a main methodology for MTL. Here we classify many regularized MTL algorithms into two main categories: learning with feature covariance and learning with task relations. The former can be viewed as a representative formulation in feature-based MTL, while the latter is for parameter-based MTL.

Objective functions in the first category can be unified as
{\small
\begin{equation}
\hspace{-0.1in}\min_{\mathbf{W},\mathbf{b},\boldsymbol{\Theta}}L(\mathbf{W},\mathbf{b})
+\frac{\lambda}{2}\mathrm{tr}(\mathbf{W}^T\boldsymbol{\Theta}^{-1}\mathbf{W})+f(\boldsymbol{\Theta}),\label{equ_first_category_obj}
\end{equation}
}\noindent
where $f(\cdot)$ denotes a regularizer or constraint on $\boldsymbol{\Theta}$. From the perspective of probabilistic modeling, the regularizer $\frac{\lambda}{2}\mathrm{tr}(\mathbf{W}^T\boldsymbol{\Theta}^{-1}\mathbf{W})$ corresponds to a matrix-variate normal distribution on $\mathbf{W}$ as $\mathbf{W}\sim\mathcal{MN}(\mathbf{0},\frac{1}{\lambda}\boldsymbol{\Theta}\otimes \mathbf{I})$. Based on this probabilistic prior, $\boldsymbol{\Theta}$ models the covariance between the features since $\frac{1}{\lambda}\boldsymbol{\Theta}$ is the row covariance matrix with each row in $\mathbf{W}$ corresponding to a feature and different tasks share the feature covariance. All the models in this category differ in the choice of the function $f(\cdot)$ on $\boldsymbol{\Theta}$. For example, methods in \cite{az05,aep08,ctly09} use $f(\cdot)$ to restrict the trace of $\boldsymbol{\Theta}$ as shown in problems (\ref{equ_MTFL_objective_2}) and (\ref{equ_cASO_objective_relax}). Moreover, multi-task feature selection methods based on the $\ell_{2,1}$ norm such as \cite{otj06,ljy09,lzx10} can be reformulated as instances of problem (\ref{equ_first_category_obj}).

Different from the first category, methods in the second category have a unified objective function as
{\small
\begin{equation}
\hspace{-0.1in}\min_{\mathbf{W},\mathbf{b},\boldsymbol{\Sigma}}L(\mathbf{W},\mathbf{b})
+\frac{\lambda}{2}\mathrm{tr}(\mathbf{W}\boldsymbol{\Sigma}^{-1}\mathbf{W}^T)+g(\boldsymbol{\Sigma}),
\label{equ_second_category_obj}
\end{equation}
}\noindent
where $g(\cdot)$ denotes a regularizer or constraint on $\boldsymbol{\Sigma}$. The regularizer $\frac{\lambda}{2}\mathrm{tr}(\mathbf{W}\boldsymbol{\Sigma}^{-1}\mathbf{W}^T)$ corresponds to a matrix-variate normal prior on $\mathbf{W}$ as
$\mathbf{W}\sim\mathcal{MN}(\mathbf{0},\mathbf{I}\otimes\frac{1}{\lambda}\boldsymbol{\Sigma})$, where $\boldsymbol{\Sigma}$ is to model the task relations since $\frac{1}{\lambda}\boldsymbol{\Sigma}$ is the column covariance with each column in $\mathbf{W}$ corresponding to a task. From this perspective, the two regularizers for $\mathbf{W}$ in problems (\ref{equ_first_category_obj}) and (\ref{equ_second_category_obj}) have different meanings even though the formulations seem a bit similar. All the methods in this category use different functions $g(\cdot)$ to learn $\boldsymbol{\Sigma}$ with different functionalities. For example, the methods in \cite{ep04,emp05,kksa07,pw10}, which utilize a priori information on task relations, directly learn $\mathbf{W}$ and $\mathbf{b}$ by defining $g(\boldsymbol{\Sigma})=0$. Some task clustering methods \cite{jbv08,zz16} identify task clusters by assuming that $\boldsymbol{\Sigma}$ has a block structure. Several task relation learning methods including \cite{zy10b,sab12,zy14,lyh16,zy17} directly learn $\boldsymbol{\Sigma}$ as a covariance matrix by constraining its trace or sparsity in $g(\boldsymbol{\Sigma})$. The trace norm regularization \cite{ptjy10} can be formulated as an instance of problem (\ref{equ_second_category_obj}).

Even though this taxonomy cannot cover all the regularized MTL methods, it can bring insights to understand regularized MTL methods better and help devise more MTL models. For example, a learning framework is proposed in \cite{zwy18} to learn a suitable multi-task model for a given multi-task problem under problem (\ref{equ_second_category_obj}) by utilizing $\bm{\Sigma}$ to represent the corresponding multi-task model.

\subsection{Other Settings in MTL}

Instead of assuming that different tasks share an identical feature representation, Zhang and Yeung \cite{zy11a} consider a multi-database face recognition problem where face recognition in a database is treated as a task. Since different face databases have different image sizes, here naturally all the tasks do not lie in the same feature space in this application, leading to a heterogeneous-feature MTL problem. To tackle this problem, a multi-task discriminant analysis (MTDA) is proposed in \cite{zy11a} by first projecting data in different tasks into a common subspace and then learning a common projection in this subspace to discriminate different classes in different tasks. In \cite{hlc12}, a latent probit model is proposed to generate data of different tasks in different feature spaces via sparse transformations on a shared latent space and then to generate labels based on this latent space.

In many MTL classification problems, each task is explicitly or implicitly assumed to be a binary classification problem as each column in the parameter matrix $\mathbf{W}$ contains model parameters for the corresponding task. It is not difficult to see that many methods in the feature learning approach, low-rank approach and decomposition approach can be directly extended to a general setting where each classification task can be a multi-class classification problem and correspondingly multiple columns in $\mathbf{W}$ contains model parameters of a multi-class classification task. Such direct extension is applicable since those methods only rely on the entire $\mathbf{W}$ or its rows but not columns as a media to share knowledge among tasks. However, to the best of our knowledge, there is no theoretical or empirical study to investigate such direct extension. For most methods in the task clustering and task relation learning approaches, such direct extension does not work since for multiple columns in $\mathbf{W}$ corresponding to one task, we do not know which one(s) can be used to represent this task. Therefore, the direct extension may not be the best solution to the general setting. In the following, we introduce four main approaches other than the direct extension to tackle the general setting in MTL where each classification task can be a multi-class classification problem. The first method is to transform the multi-class classification problem in each task into a binary classification problem. For example, multi-task metric learning \cite{pw10,yhl13} can do that by treating a pair of data points from the same class as positive and that from different classes as negative. The second recipe is to utilize the characteristics of learners. For example, the linear discriminant analysis can handle binary and multi-class classification problems in a unified formulation and hence MTDA \cite{zy11a} can naturally handle them without changing the formulation. The third approach is to directly learn label correspondence among different tasks. In \cite{qscvp10}, two learning tasks, which share the training data, aim to maximize the mutual information to identify the correspondence between labels in different tasks. By assuming that all the tasks share the same label space, the last approach including \cite{yh17a,yh17b,lcwy17} organizes the model parameters of all the tasks in a tensor where the model parameters of each task form a slice. Then the parameter tensor can be regularized by tensor trace norms \cite{yh17b} and a tensor-variate normal prior \cite{lcwy17}, or factorized as a product of several low-rank matrices or tensors \cite{yh17a}.

Most MTL methods assume that the training data in each task are stored in a data matrix. In some case, the training data in each task exhibit a multi-modal structure and hence they are represented in a tensor instead of a matrix. Multilinear multi-task methods proposed in \cite{rabp13,wst14} can handle this situation by employing tensor trace norms as a generalization of the trace norm to perform the regularization.

\subsection{Optimization Techniques in MTL}

Optimization techniques used in MTL can be categorized into three main classes as follows.
\begin{itemize}

\item Gradient descent method and its variants: The gradient descent method can be used to optimize smooth unconstrained objective functions possessed by many MTL models. If the unconstrained objective function is non-smooth, the subgradient can be used instead and then the gradient descent method can also be used. When there are some constraints in the objective function of MTL models \cite{ptjy10,bc15}, the projected gradient descent method can be used to project the updated solution in each step to the space defined by constraints. For deep MTL models, stochastic gradient descent methods can be used. Moreover, the GradNorm \cite{sk18} is devised to normalize gradients to balance the learning of multiple tasks and \cite{ykglhf20} proposes the gradient surgery to avoid the interference between task gradients. Differently, \cite{cblr18} studies MTL from the perspective of multi-objective optimization by learning dynamic loss weights.

\item Block Coordinate Descent (BCD) method: The parameters in many MTL models can be divided into several blocks. For example, parameters in learning functions of all the tasks form a block and parameters to represent task relations are from another block. Directly optimizing the objective function of such a MTL model with respect to parameters in all blocks together is not easy. The BCD method, which is also known as the alternating method, is widely used in the MTL literature, e.g., \cite{az05,aep08,jbv08,ctly09,zy10b,kgs11,ls12,kd12,mpr13,zhang13,zy14,zz16,lyh16}, to alternatively optimize each block of parameters while fixing parameters in other blocks. Hence, each step of the BCD method will solve several subproblems, each of which is to optimize with respect to a block of parameters. Compared with the original objective function, each subproblem is easier to be solved and so the BCD method can help reduce the optimization complexity.

\item Proximal method \cite{pb14}: For a nonsmooth objective function, which is the sum of smooth and nonsmooth functions, in an MTL model, the proximal method is frequently used (e.g., \cite{ljy09,cly10,czy11,gyz12,zk12,zw13,gzfy14,hz15a,hz15b,zsyclr15,lwyr16,zsyclr17}) to construct a proximal problem by replacing the smooth function with a quadratic function that may be constructed based on its Taylor series in various ways and the resulting proximal problem is usually easier to be solved than the original problem. The proximal method can accelerate the convergence rate of the optimization process or facilitate the design of distributed optimization algorithms.

\end{itemize}

\section{MTL with Other Learning Paradigms}
\label{sec:combination_other_paradigm}

In the previous section, we review different MTL approaches for supervised learning tasks. In this section, we overview some works on the combination of MTL with other learning paradigms in machine learning, including unsupervised learning such as clustering, semi-supervised learning, active learning, reinforcement learning, multi-view learning and graphical models, to either improve the performance of supervised MTL further via additional information such as unlabeled data or use MTL to help improve the performance of other learning paradigms.

In most applications, labeled data are expensive to collect but unlabeled data are abundant. So in some MTL applications, the training dataset of each task consists of both labeled and unlabeled data, hence we hope to exploit useful information contained in the unlabeled data to further improve the performance of supervised learning tasks. In machine learning, semi-supervised learning and active learning are two ways to utilize unlabeled data but in different ways. Semi-supervised learning aims to exploit geometrical information contained in the unlabeled data, while active learning selects representative unlabeled data to query an oracle with the hope of increasing the labeling cost as little as possible. Hence semi-supervised learning and active learning can be combined with MTL, leading to three new learning paradigms including semi-supervised multi-task learning \cite{llc07,lllsc09,zy09}, multi-task active learning \cite{rthr08,amg14,ft15} and semi-supervised multi-task active learning \cite{llc09a}. Specifically, a semi-supervised multi-task classification model is proposed in \cite{llc07,lllsc09} to use random walk to exploit unlabeled data in each task and then cluster multiple tasks via a relaxed Dirichlet process. In \cite{zy09}, a semi-supervised multi-task Gaussian process for regression tasks, where different tasks are related via the hyperprior on the kernel parameters in Gaussian processes of all the tasks, is proposed to incorporate unlabeled data into the design of the kernel function in each task to achieve the smoothness in the corresponding functional spaces. Different from these semi-supervised multi-task methods, multi-task active learning adaptively selects informative unlabeled data for multi-task learners and hence the selection criterion is the core research issue. Reichart et al. \cite{rthr08} believe that data instances to be selected should be as informative as possible for a set of tasks instead of only one task and hence they propose two protocols for multi-task active learning. In \cite{amg14}, the expected error reduction is used as a criterion where each task is modeled by a supervised latent Dirichlet allocation model. Inspired by multi-armed bandits which balance the trade-off between the exploitation and exploration, a selection strategy is proposed in \cite{ft15} to consider both the risk of a multi-task learner based on the trace norm regularization and the corresponding confidence bound. In \cite{lz16}, the MTRL method (i.e., problem (\ref{equ_MTRL_objective})) is extended to the interactive setting where a human expert is enquired about partial orderings of pairwise task covariances based an inconsistency criterion. In \cite{pl17}, a proposed generalization bound is used to select a subset from multiple unlabeled tasks to acquire labels to improve the generalization performance of all the tasks. For semi-supervised multi-task active learning, Li et al. \cite{llc09a} propose a  model to use the Fisher information as a criterion to select unlabeled data to acquire their labels with the semi-supervised multi-task classification model \cite{llc07,lllsc09} as the classifier for each task.

MTL achieves the performance improvement in not only supervised learning tasks but also unsupervised learning tasks such as clustering. In \cite{zz10}, a multi-task Bregman clustering method is proposed based on single-task Bregman clustering by using the earth mover distance to minimize distances between any pair of tasks in terms of cluster centers and then in \cite{zz13,zzl15a}, an improved version of \cite{zz10} and its kernel extension are proposed to avoid the negative effect caused by the regularizer in \cite{zz10} via choosing the better one between single-task and multi-task Bregman clustering. In \cite{glh11}, a multi-task kernel $k$-means method is proposed by learning the kernel matrix via both MMD between any pair of tasks and the Laplacian regularization that helps identify a smooth kernel space. In \cite{zhang15c}, two proposed multi-task clustering methods are extensions of the MTFL and MTRL methods by treating labels as cluster indicators to be learned. In \cite{wwlcw15}, the principle of MTL is incorporated into the subspace clustering by capturing correlations between data instances. In \cite{zzl16}, a multi-task clustering method belonging to instance-based MTL is proposed to share data instances among different tasks. In \cite{ymyns15}, a multi-task spectral clustering algorithm, which can handle the out-of-sample issue via a linear function to learn the cluster assignment, is proposed to achieve the feature selection among tasks via the $\ell_{2,1}$ regularization \cite{zlzjw17}. \cite{zzll18b} proposes to identify the task cluster structure and learn task relations together.

Reinforcement Learning (RL) is a promising area in machine learning and has shown superior performance in many applications such as game playing (e.g., Atari and Go) and robotics. MTL can help boost the performance of reinforcement learning, leading to Multi-task Reinforcement Learning (MRL). Some works \cite{wfrt07,llc09b,lg10,clr14,akl17,dds17,ser17,bbrw19,vnnbkttl19,ignsbw20} adapt the ideas introduced in Section \ref{sec:category} to MRL.
Specifically, in \cite{wfrt07} where a task solves a sequence of Markov Decision Processes (MDPs), a hierarchical Bayesian infinite mixture model is used to model the distribution over MDPs and for each new MDP, previously learned distributions are used as an informative prior.
In \cite{llc09b}, a regionalized policy representation is introduced to characterize the behavior of an agent in each task and a Dirichlet process is placed over regionalized policy representations across multiple tasks to cluster tasks.
In \cite{lg10}, a Gaussian process temporal-difference value function model is used for each task and a hierarchical Bayesian approach is to model the distribution over value functions in different tasks.
Calandriello et al. \cite{clr14} assume that parameter vectors of value functions in different tasks are jointly sparse and then extend the MTFS method with the $\ell_{2,1}$ regularization as well as the MTFL method to learn value functions in multiple tasks together.
In \cite{akl17}, a model associating each subtask with a modular subpolicy is proposed to learn from policy sketches, which annotate tasks with sequences of named subtasks and provide information about high-level structural relationships among tasks.
In \cite{dds17}, a multi-task contextual bandit is introduced to leverage or learn similarities in contexts among arms to improve the prediction of rewards from contexts.
In \cite{ser17}, a multi-task linearly solvable MDP, whose task basis matrix contains a library of component tasks shared by all the tasks, is proposed to maintain a parallel distributed representation of tasks each of which enables an agent to draw on macro actions simultaneously.
In \cite{bbrw19}, a multi-task deep RL model based on the attention can automatically group tasks into sub-networks on a state-level granularity.
In \cite{vnnbkttl19}, a sharing experience framework is introduced to use task-specific rewards to identify similar parts defined as shared-regions which can guide the experience sharing of task policies.
In \cite{ignsbw20}, multi-task soft option learning, a hierarchical framework based on planning as inference, is regularized by a shared prior to avoid training instabilities and allow the fine-tuning of options for new tasks without forgetting learned policies.
The idea of compression and distillation have been incorporated into MRL as in \cite{pbs16,rcgdkpmkh16,opahv17,tbcqkhhp17}.
For example, in \cite{pbs16}, the proposed Actor-Mimic method combines both deep reinforcement learning and model compression techniques to train a policy network which can learn to act for multiple tasks.
In \cite{rcgdkpmkh16}, a policy distillation method is proposed to not only train an efficient network to learn the policy of an agent but also consolidate multiple task-specific policies into a single policy.
In \cite{opahv17}, the problem of multi-task multi-agent reinforcement learning under the partial observability is addressed by distilling decentralized single-task policies into a unified policy across multiple tasks.
In \cite{tbcqkhhp17}, each task has its own policy which is constrained to be close to a shared policy that is trained by the distillation.
Some works \cite{bbt17,sr17,sjhr18,esmsmwdf18,tkb18,lbkh19} in MRL focus on online and distributed settings.
Specifically, in \cite{bbt17}, a distributed MRL framework is devised to model it as an instance of general consensus and an efficient decentralized solver is developed.
In \cite{sr17,sjhr18}, multiple goal-directed tasks are learned in an online setup without the need for expert supervision by actively sampling harder tasks.
In \cite{esmsmwdf18}, a distributed agent is developed to not only use resources more efficiently in single-machine training but also scale to thousands of machines without sacrificing data efficiency or resource utilization.
\cite{tkb18} formulates MRL from a perspective of variational inference and it proposes a novel distributed solver with quadratic convergence guarantees.
In \cite{lbkh19}, an online learning algorithm is proposed to dynamically combine different auxiliary tasks which provide gradient directions to speed up the training of the main reinforcement learning task.
Some works study the theoretical foundation of MRL.
For example, in \cite{dtbrp20}, sharing representations among tasks is analyzed with theoretical guarantees to highlight conditions to share representations and finite-time bounds of approximated value-iteration are extended to the multi-task setting.
Moreover, there are some works to design novel MRL methods. For example, in \cite{sxs18}, a MRL framework is proposed to train agent to employ hierarchical policies that decide when to use a previously learned policy and when to learn a new skill with a temporal grammar that helps the agent learn complex temporal dependencies.
\cite{hsecsh19} studies the problem of parallel learning of multiple sequential-decision tasks and proposes to automatically adapt the contribution of each task to the updates of the agent to make all tasks have comparable impacts on the learning dynamics.
In \cite{gpapagm20}, a self-supervised representation learning algorithm is proposed for multi-task deep RL to capture structured information about environment dynamics based on multi-step predictive representations of future observations.

Multi-view learning assumes that each data point is associated with multiple sets of features where each set corresponds to a view and it usually exploits information contained in multiple views for supervised or semi-supervised learning tasks. Multi-task multi-view learning extends multi-view learning to the MTL setting where each task is a multi-view learning problem. Specifically, in \cite{hl11}, a graph-based method is proposed for multi-task multi-view classification problems. In a task, each view is enforced to be consistent with both other views and labels, while different tasks are expected to have similar predictions on views they share, making views as a bridge to construct the task relatedness. In \cite{zh12}, both a regularized MTL method \cite{emp05} and the MTRL method are applied to each view of different tasks and different views in a task are expected to achieve an agreement on unlabeled data. Different from \cite{hl11,zh12} which study the multi-task multi-view classification problem, in \cite{zzl15b,zzll16}, two multi-task multi-view clustering methods are proposed and both methods consider three factors: within-view-task clustering which conducts clustering on each view in a task, view relation learning which minimizes the disagreement among views in a task, and low-rank structure learning which aims to learn a shared subspace for different tasks under a common view. The difference between these two methods is that the first method uses a bipartite graph co-clustering method for nonnegative data while the other one adopts a semi-nonnegative matrix tri-factorization to cluster general data. In \cite{zlz16}, a multi-label multi-view algorithm is proposed to not only learn common features via the $\ell_{2,1}$ regularization but also identify useless views via the Frobenius norm. In multi-task multi-view learning, each task is usually supplied with both labeled and unlabeled data, hence this paradigm can also be viewed as another way to utilize unlabeled information for MTL. A deep multi-task multi-view model is proposed in \cite{zch19} to fuse all the views based on the cross-stitch network.

MTL can help learn more accurate structure in graphical models. In \cite{nc07}, an algorithm is proposed to learn Bayes network structures by assuming that different networks/tasks share similar structures via a common prior and then a heuristic search is used to find structures with high scores for all the tasks. With a similar idea, multiple Gaussian graphical models are jointly learned in \cite{hs10} by assuming joint sparsity among precision matrices via the $\ell_{\infty,1}$ norm regularization. In \cite{ol12}, some domain knowledge about task relations is incorporated into the learning of multiple Bayesian networks. By viewing the feature interaction matrix as a form of graphical models to model pairwise relations between features, two models are proposed in \cite{lxbjz16} to learn a quadratical function, where the feature interaction matrix defines the quadratic term, for each task based on the $\ell_{2,1}$ and tensor trace norm regularization, respectively.

According to the above discussions, we can see that most research works discussed in this section follow the spirits of MTL approaches introduced in Section \ref{sec:category} and adapt to their own settings.

\section{Handling Big Data}
\label{sec:big_data}

When the number of tasks is large, the total number of training data in all the tasks can be very big and hence a `big' aspect in MTL denotes the number of tasks. In this case, we can either devise online, parallel, or distributed MTL models to accelerate the learning process. Another `big' aspect in MTL lies in the data dimensionality which can be very high. In this situation, we can speedup the learning via feature selection, dimensionality reduction and feature hashing to reduce the dimension without losing too much useful information. In this section, we review some relevant works.

When the number of tasks is very big, we can devise some parallel MTL methods to speedup the learning process on multi-CPU or multi-GPU devices. As a representative formulation in feature-based MTL, problem (\ref{equ_first_category_obj}) is easy to parallelize since when given the feature covariance matrix $\bm{\Theta}$, the learning of different tasks can be decoupled. However, for problem (\ref{equ_second_category_obj}) in parameter-based MTL, the situation is totally different since even given the task covariance matrix $\bm{\Sigma}$, different tasks are still coupled, making the direct parallelization fail. In order to parallelize problem (\ref{equ_second_category_obj}), Zhang \cite{zhang15b} uses the FISTA algorithm to design a surrogate function for problem (\ref{equ_second_category_obj}) with a given $\bm{\Sigma}$, where the surrogate function is decomposable with respect to tasks, leading to a parallel design for MTL based on different loss functions including the hinge, $\epsilon$-insensitive and square losses.

Online multi-task learning is also capable of handling a big number of tasks. In \cite{dls06,dls07}, under a setting where all the tasks contribute toward a common goal, the relation between tasks is measured via a global loss function and several online algorithms are proposed to use absolute norms as the global loss function. In \cite{lps09}, online MTL algorithms are devised by modeling the task relatedness via hard constraints that the $m$-tuple of actions for tasks satisfies. In \cite{ccg10}, perceptron-based online algorithms are proposed for multi-task binary classification problems where task similarities are measured based on either the geometric closeness of the task reference vectors or the dimension of their spanned subspace. In \cite{pdn10}, a recursive Bayesian online algorithm based on Gaussian processes is devised to update both estimations and confidence intervals when data instances arrive sequentially. In \cite{srdv11}, an online version of the MTRL method is proposed to update both the model parameters and task covariance in a sequential way.
An online multi-task learning algorithm is proposed in \cite{mlcy16} to jointly learn the per-task model and the task relations by smoothing the loss function of each task w.r.t. a task distribution and adaptively refining this distribution over time. In \cite{yzg17}, an online multi-task model is proposed to learn both a low-rank component and a group sparse component to characterize task relations. In \cite{hzlhm17}, a multi-task passive-aggressive method is proposed to learn multiple relative similarity learning tasks, each of which is to learn a similarity function from data with relative constraints.
In \cite{yzzg19}, a Gaussian distribution, whose mean or covariance consists of a local component for each task and a global component shared by all the tasks, over each task is used as a confidence measure to guide the online MTL process.

Training data can locate at different devices, making the design of distributed MTL models important. In \cite{wks16}, a communication-efficient distributed MTL algorithm, where each machine learns a task, based on the debiased Lasso is proposed to learn jointly \mbox{sparse} features in a high-dimensional space. In \cite{lph17}, the MTRL method (i.e., problem (\ref{equ_MTRL_objective})) is extended to the distributed setting based on a stochastic dual coordinate ascent method. In \cite{xblz17}, to protect the privacy of data, a privacy-preserving distributed MTL method is proposed based on a privacy-preserving proximal gradient algorithm with asynchronous updates. In \cite{scst17}, federated multi-task learning is proposed as an extension of distributed multi-task learning to consider both stragglers and fault tolerance. In \cite{zzhslmh18}, a distributed multi-task algorithm is proposed under the online MTL setting.

For high-dimensional data in MTL, we can use multi-task feature selection methods to reduce the dimension or extend single-task dimension reduction techniques to the multi-task setting as did in \cite{zy11a}. Another option is to use the feature hashing and in \cite{wdlsa09}, multiple hashing functions are proposed to accelerate the joint learning of multiple tasks.

\begin{table*}[!htb]
\caption{The classification of works about MTL applications in different areas according to different MTL approaches.} \label{table-applications}\centering
\resizebox{0.85\textwidth}{!}{
\begin{tabular}{|c|l|l|l|l|l|}
\hline
\multirow{2}{*}{Approach} & \multirow{2}{*}{Computer Vision} & Bioinformatics \&  & \multirow{2}{*}{Speech \& NLP} & \multirow{2}{*}{Web} & \multirow{2}{*}{Miscellaneous}\\
                 &                & Health Informatics &                     &                      &                          \\
\hline
\multirow{10}{*}{Feature Transformation}
& Visual tracking \cite{zgla12,zgla13}    & Protein subcellular location prediction \cite{xpxy11} & Speech synthesis \cite{wvwk15,hwrysm15} & Learning to rank \cite{bzxzstzc09} & Stock prediction \cite{gb96} \\
& Action recognition \cite{yhtyw13}       & Protein interaction prediction \cite{qtckw10}     & Speech recognition \cite{br15,cweh15}   &                                    & Localization \cite{zpyp08} \\
& Facial landmark detection \cite{zllt14} & Biological image classification \cite{zlzskyj15} & Jointly learning of NLP tasks \cite{cw08}   &                                    &              \\
& Scene classification \cite{lsh14}       &                                & Dialog state tracking \cite{mstgsvwy15} &                              &                 \\
& Attribute prediction \cite{awlj15}      &                              & Machine translation \cite{llsvk16}      &                                    &                 \\
& Image rotation \cite{yjycpk15}          &                              & Syntactic parsing \cite{llsvk16}        &                                    &                 \\
& Immediacy prediction \cite{coyw15}      &                              &                                         &                                    &                 \\
& Pose estimation \cite{llc15}            &                              &                                         &                                    &                 \\
& Thumbnail selection \cite{lmzcl15}      &                              &                                         &                                    &                 \\
& Face verification \cite{wzz09}          &                              &                                         &                                    &                 \\
\hline
\multirow{7}{*}{Feature Selection}
& Face and object recognition \cite{yy10} & siRNA efficacy prediction \cite{lxzxcy10} & Microblog analysis \cite{zsyclr15} & Behavioral targeting \cite{aadsa12} &           \\
& Brain imaging \cite{wnhrdss11} & Genetic marker detection \cite{pkx10}     &                                    &                                     &           \\
&                                & Mental state examination \cite{zyly11}    &                                    &                                     &           \\
&                                & Predict cognitive outcome \cite{wzylrfkrss12}&                                 &                                     &           \\
&                                & Survival analysis \cite{lwyr16}           &                                    &                                     &           \\
&                                & Genetic trait prediction \cite{hkp16}     &                                    &                                     &           \\
&                                & Gene expression association \cite{zgp10}  &                                    &                                     &           \\
\hline
\multirow{2}{*}{Low-Rank}
& Image segmentation \cite{clwhy11} & Identification of longitudinal       &                      &                          & Climate prediction \cite{xtlz16} \\
& Saliency detection\cite{llyy12}   & phenotypic markers \cite{wnhykrss12} &                      &                          &                                  \\
\hline
\multirow{5}{*}{Task Clustering}
& Image segmentation \cite{awswcd08} & Brain-computer interfaces \cite{aga10} &                      &                          &                      \\
& Age estimation \cite{zy10c}        & Personalized medical                   &                      &                          &                      \\
& Facial action unit                 & treatment \cite{xzt15}                 &                      &                          &                      \\
& prediction \cite{amv15}            & Genetic trait prediction \cite{hkp16}  &                      &                          &                      \\
& Action recognition \cite{lsnk17}   &                                        &                      &                          &                      \\
\hline
\multirow{7}{*}{Task Relation Learning}
&  & Protein subcellular location prediction \cite{xpxy11}& Sentiment classification \cite{wh15} & Web search ranking \cite{csvwzt10}   & Localization \cite{zpyp08} \\
&  & Organism modeling \cite{wlar10}        &                                      & Collaborative filtering \cite{zcy10} & Robotics \cite{cwkv08,yz09} \\
&  & MHC-I binding prediction \cite{wtar10} &                                      & Display advertising \cite{ads14}     & Trajectory regression \cite{zn13} \\
&  & Splice-site prediction \cite{wtar10}   &                                      &                                      & Traffic sign recognition \cite{lwzzl17} \\
&  & Prioritization of disease genes \cite{mv11} &                                      &                                      & Soil moisture forecasts \cite{xtzl17} \\
&  & Protein interaction prediction \cite{kck13} &                                      &                                      &                      \\
&  & Identifying antigenic variants \cite{hlwzzw19} &                                 &                                      &                      \\
\hline
\multirow{3}{*}{Decomposition}
& Multi-view tracking \cite{hmpt13}        & Genetic trait prediction \cite{hkp16}        &                      &                          &                      \\
& Pose estimation \cite{yrrls13}           & Protein interaction prediction \cite{kmck17} &                      &                          &                      \\
& Person re-identification \cite{syztdg15} &                                              &                      &                          &                      \\
\hline
\end{tabular}
}
\end{table*}

\section{Applications}
\label{sec:applications}

MTL has many applications in various areas including computer vision, bioinformatics, health informatics, speech, NLP, web, and so on. In Table \ref{table-applications}, we categorize different MTL problems in each application area according to MTL approaches they used, where the classification of MTL approaches has been already introduced in Section \ref{sec:category}. In the last column of Table \ref{table-applications}, we list some problems in various application areas which are different from other columns. For application problems listed in Table \ref{table-applications}, application-dependent MTL models have been proposed to solve them.\footnote{For details of those models, refer to an arXiv version \cite{zy17b} of this paper.} Though these models are different from each other, there are some characteristics in respective areas. For example, in computer vision, deep MTL models, most of which belong to the feature transformation approach, exhibit good performance, making this approach popular in computer vision. In bioinformatics and health informatics, the interpretability of learning models is more important in some sense. Therefore, the feature selection and task relation learning approaches are widely used in this area as the former approach can identify useful features and the latter one can quantitatively show task relations. In speech and NLP, the data exhibit a sequential structure, which makes recurrent-neural-network-based deep MTL models in the feature transformation approach play an important role. As the data in web applications is of a large scale, this area favors simple models such as linear models or their ensembles based on boosting. Among all the MTL approaches, the feature transformation, feature selection and task relation learning approaches are among the most widely used MTL approaches in different application areas according to Table \ref{table-applications}.

When encountering a new application problem which can be modeled as a MTL problem, we need to judge whether tasks in this problem are related in terms of either low-level features or high-level concepts. If so, by treating Table \ref{table-applications} as a look-up table. we can identify a problem in Table \ref{table-applications} similar to the new problem and then adapt the corresponding MTL model to solve the new problem. Otherwise, we can try popular MTL approaches in the respective area.

\section{Theoretical Analyses}
\label{sec:analysis}

As well as designing MTL models and exploiting MTL applications, there are some works to study theoretical aspects of MTL and here we review them.

The generalization bound, which is to upper-bound the generalization loss in terms of the training loss, model complexity and confidence, is core in learning theory since it can identify the learnability and induce the sample complexity. There are several works \cite{baxter95,baxter00,az05,maurer06a,maurer06b,juba06,bdb08,cm12,kst12,pm13,pb15,zhang15a,mpr16,ylkma18} to study the generalization bound of different MTL models. In Table \ref{table-analyses}, we compare those works in terms of the analyzed MTL model, analysis tool and the convergence rate of the corresponding bound which is based on the number of tasks (i.e., $m$) and the average number of data points per task (i.e., $n_0$). According to Table \ref{table-analyses}, we can see that some works (i.e., \cite{baxter95,baxter00,juba06,bdb08,cm12,pb15,zhang15a}) analyze different MTL models based on various analysis tools and the best convergence rate is $O(\frac{1}{\sqrt{mn_0}})$. Though MTL models analyzed in \cite{az05,maurer06a,mpr13,mpr16} are not the same, those MTL models exhibit similar objectives to learn a linear or nonlinear feature transformation shared by all the tasks, and the convergence rates are $O(\frac{1}{\sqrt{mn_0}})$ except \cite{mpr13}. The trace norm regularization (i.e., Problem (\ref{equ_TNR_objective})) is analyzed in \cite{zhang15a,kst12,pm13,ylkma18}, among which \cite{ylkma18} has the best convergence rate based on the local Rademacher complexity. A related MTL model based on the Schatten norm regularization is analyzed in \cite{maurer06b} with an $O(\frac{1}{\sqrt{mn_0}})$ convergence rate. For the graph regularization \cite{ep04,emp05} analyzed in \cite{maurer06b,ylkma18}, the local Rademacher complexity leads to a better convergence rate (i.e., $O(\frac{1}{(mn_0)^\alpha})$ for some constant $\alpha\in(0.5,1)$) and a similar observation holds for problem (\ref{equ_MTFS_l21_objective}). In a word, various analysis tools can be used to analyze MTL models and among them, the local Rademacher complexity can derive tighter generalization bounds than others.

Besides generalization bounds, there are some works to study other theoretical problems in MTL. For example, Argyriou et al. \cite{amp09,amp10} discuss conditions where representer theorems hold for regularized MTL algorithms. Several studies \cite{lptv09,owj11,klw11} investigate conditions to well recover true features for multi-task feature selection models.

\begin{table}[!htb]%\small
\caption{Comparison of generalization bounds derived in different works in terms of the analyzed MTL model, analysis tool and the convergence rate of the bound. $m$ denotes the number of tasks and $n_0$ denotes the average number of data points per task. } \label{table-analyses}\centering
\resizebox{0.5\textwidth}{!}{
\begin{tabular}{|c|c|c|c|}
\hline
MTL Model                             & Reference                & Analysis Tool                   & Convergence rate  \\
\hline
Tasks from an environment             &	\cite{baxter95,baxter00} & VC dimension \& Covering number & $O(1/\sqrt{mn_0})$ \\
\hline
Task distributions can be transformed &	\cite{bdb08}             & VC dimension	                   & $O(1/\sqrt{mn_0})$ \\
\hline
Task clustering	                      & \cite{cm12}              & VC dimension	                   & $O(m\ln(n_0/m)/n_0)$ \\
\hline
Multi-task kernel classifier          &	\cite{pb15}       	     & Covering number	               & $O(1/\sqrt{mn_0})$ \\
\hline
Problem~(\ref{equ_DM_objective}) with Eq.~(\ref{equ_RTRN1_U_V_formulation}) & \cite{zhang15a} &	Multi-task stability & $O(m\sqrt{m}/n_0)$ \\
\hline
Multi-task data compression           &	\cite{juba06}            & Kolmogorov complexity	       & $O(1/\sqrt{mn_0})$ \\
\hline
Problem (\ref{equ_ASO_objective})	  & \cite{az05}              & Covering number	               & $O(1/\sqrt{mn_0})$ \\
Problem (\ref{equ_ASO_objective}) without $\mathbf{U}$ & \cite{maurer06a} & Rademacher complexity  & $O(1/\sqrt{mn_0})$ \\
Problem (\ref{equ_MTSC_objective})    & \cite{mpr13}             & Rademacher complexity           & $O(1/\sqrt{n_0})$ \\
Learn a common feature transformation &	\cite{mpr16}             & Gaussian average	               & $O(1/\sqrt{mn_0})$ \\
\hline
\multirow{4}{*}{Problem (\ref{equ_TNR_objective})} & \cite{zhang15a} & Multi-task stability	       & $O(m/n_0)$ \\
\cline{2-4}
                                      & \cite{kst12}             & Rademacher complexity	       & $O(\ln(m)/\sqrt{n_0})$\\
\cline{2-4}
                                      & \cite{pm13}              & Rademacher complexity           & $O(\max(\sqrt{\ln(mn_0)/mn_0},1/\sqrt{n_0}))$ \\
\cline{2-4}
                                      & \cite{ylkma18}           & Local Rademacher complexity     & $O(1/(mn_0)^\alpha)$, $0.5<\alpha<1$ \\
\hline
Schatten-norm-regularized MTL models  & \cite{maurer06b}         & Rademacher complexity           & $O(1/\sqrt{mn_0})$ \\
\hline
\multirow{2}{*}{Graph regularizers \cite{ep04,emp05}} & \cite{maurer06b} & Rademacher complexity   & $O(1/\sqrt{mn_0})$ \\
\cline{2-4}
                                      & \cite{ylkma18}           & Local Rademacher complexity     & $O(1/(mn_0)^\alpha)$, $0.5<\alpha<1$ \\
\hline
\multirow{2}{*}{Problem (\ref{equ_MTFS_l21_objective})} &  \cite{kst12} & Rademacher complexity    & $O(1/\sqrt{n_0})$ \\
\cline{2-4}
                                      & \cite{ylkma18}           & Local Rademacher complexity     & $O(1/(mn_0)^\alpha)$, $0.5<\alpha<1$ \\
\hline
\end{tabular}
}
\end{table}

\section{Conclusions and Discussions}
\label{sec:conclusions}

In this paper, we survey different aspects of MTL. First, after giving the definition of MTL, we give a classification of supervised MTL models into five main approaches and discuss their characteristics. Then we review the combinations of MTL with other learning paradigms. The online, parallel and distributed MTL models as well as dimensionality reduction and feature hashing are discussed to speedup the learning process. The applications of MTL in various areas are introduced to show the usefulness of MTL and theoretical aspects of MTL are discussed.

In future studies, there are several issues to be addressed. Firstly, outlier tasks, which are unrelated to other tasks, are well known to hamper the performance of all the tasks when learning them jointly. There are some methods to alleviate negative effects that outlier tasks bring. However, there lacks principled ways and theoretical analyses to study the resulting negative effects. In order to make MTL safe to be used by human, this is an important issue and needs more studies.

Secondly, deep learning has become a dominant approach in many areas and several multi-task deep models belonging to the feature transformation, low-rank, task clustering and task relation learning approaches have been proposed as reviewed in Sections \ref{sec:category}, \ref{sec:combination_other_paradigm} and \ref{sec:applications}. As discussed, most of them only share hidden layers. This approach is powerful when all the tasks are related, but it is vulnerable to noisy and outlier tasks that can deteriorate the performance dramatically. We believe that it is desirable to design flexible and robust deep multi-task models.

Lastly, existing studies mainly focus on supervised learning tasks, and only a few ones are on other tasks such as unsupervised learning, semi-supervised learning, active learning, multi-view learning and reinforcement learning tasks. It is natural to adapt or extend the five approaches introduced in Section \ref{sec:category} to those non-supervised learning tasks. We think that such adaptation and extension require more efforts to design appropriate models. Moreover, it is worth trying to apply MTL to other areas in artificial intelligence such as logic and planning to broaden its application scopes.

%%% use section* for acknowledgment
%\ifCLASSOPTIONcompsoc
%  % The Computer Society usually uses the plural form
%  \section*{Acknowledgments}
%\else
%  % regular IEEE prefers the singular form
%  \section*{Acknowledgment}
%\fi

\vspace{0.1in}

%We thank anonymous reviewers for their constructive comments to improve this work.
\noindent{\bf Acknowledgments} This work is supported by NSFC 62076118.

% Can use something like this to put references on a page
% by themselves when using endfloat and the captionsoff option.
\ifCLASSOPTIONcaptionsoff
  \newpage
\fi

% trigger a \newpage just before the given reference
% number - used to balance the columns on the last page
% adjust value as needed - may need to be readjusted if
% the document is modified later
%\IEEEtriggeratref{8}
% The "triggered" command can be changed if desired:
%\IEEEtriggercmd{\enlargethispage{-5in}}

% references section

% can use a bibliography generated by BibTeX as a .bbl file
% BibTeX documentation can be easily obtained at:
% http://www.ctan.org/tex-archive/biblio/bibtex/contrib/doc/
% The IEEEtran BibTeX style support page is at:
% http://www.michaelshell.org/tex/ieeetran/bibtex/
%\bibliographystyle{IEEEtran}
% argument is your BibTeX string definitions and bibliography database(s)
%\bibliography{IEEEabrv,../bib/paper}
%
% <OR> manually copy in the resultant .bbl file
% set second argument of \begin to the number of references
% (used to reserve space for the reference number labels box)
%\begin{thebibliography}{1}
%
%\bibitem{IEEEhowto:kopka}
%H.~Kopka and P.~W. Daly, \emph{A Guide to {\LaTeX}}, 3rd~ed.\hskip 1em plus
%  0.5em minus 0.4em\relax Harlow, England: Addison-Wesley, 1999.
%
%\end{thebibliography}

\bibliographystyle{ieeetr}
\bibliography{MTL_Survey}

\end{document}